\documentclass[times,twocolumn,final]{elsarticle}

\usepackage{medima}
\usepackage{framed,multirow}

\usepackage{amssymb}
\usepackage{latexsym}
\usepackage{booktabs}
\usepackage{url}
\usepackage{xcolor}

\usepackage{hyperref}
\usepackage{graphicx}
\definecolor{newcolor}{rgb}{.8,.349,.1}

\usepackage{amsmath}
\usepackage{cases}

\usepackage{amssymb}

\usepackage{amsfonts} 
\usepackage{amsbsy}   
\usepackage{bbm}      
\DeclareUnicodeCharacter{FF08}{(} 

\DeclareUnicodeCharacter{FF09}{)} 

\journal{Medical Image Analysis}

\begin{document}

\verso{Given-name Surname \textit{et~al.}}

\begin{frontmatter}

 \title{$C^2$M-DoT: Cross-modal consistent multi-view medical report generation with domain transfer network}

\author[1]{Ruizhi \snm{Wang}}
\author[1]{Xiangtao \snm{Wang}}
\author[1]{Jie \snm{Zhou}}
\author[3,4]{Thomas \snm{Lukasiewicz}}
\author[1]{Zhenghua \snm{Xu}\corref{cor1}}
\cortext[cor1]{Corresponding author, email: zhenghua.xu@hebut.edu.cn.}


\address[1]{State Key Laboratory of Reliability and Intelligence of Electrical Equipment, School of Health Sciences and Biomedical Engineering, Hebei University of Technology, Tianjin, China}
\address[2]{College of Computer Science and Software Engineering and Guangdong Laboratory
of Artificial Intelligence and Digital Economy (SZ), Shenzhen University, Shenzhen, China}
\address[3]{Institute of Logic and Computation, Vienna University of Technology, Vienna, Austria}
\address[4]{Department of Computer Science, University of Oxford, Oxford, United Kingdom}


\begin{abstract}
In clinical scenarios, multiple medical images with different views are usually generated simultaneously, and these images have high semantic consistency. However, most existing medical report generation methods only consider single-view data. The rich multi-view mutual information of medical images can help generate more accurate reports, however, the dependence of multi-view models on multi-view data in the inference stage severely limits their application in clinical practice. In addition, word-level optimization based on numbers ignores the semantics of reports and medical images, and the generated reports often cannot achieve good performance. Therefore, we propose a cross-modal consistent multi-view medical report generation with a domain transfer network ($C^2$M-DoT). Specifically, (i) a semantic-based multi-view contrastive learning medical report generation framework is adopted to utilize cross-view information to learn the semantic representation of lesions; (ii) a domain transfer network is further proposed to ensure that the multi-view report generation model can still achieve good inference performance under single-view input; (iii) meanwhile, optimization using a cross-modal consistency loss facilitates the generation of textual reports that are semantically consistent with medical images. Extensive experimental studies on two public benchmark datasets demonstrate that $C^2$M-DoT substantially outperforms state-of-the-art baselines in all metrics. Ablation studies also confirmed the validity and necessity of each component in $C^2$M-DoT.
\end{abstract}

\begin{keyword}
\KWD Multi-view contrastive learning\sep Domain transfer\sep Cross-modal consistency\sep Medical report generation\sep Chest X-Ray
\end{keyword}

\end{frontmatter}


\section{Introduction}
\label{sec1}

With the rapid development of medical imaging technology, medical imaging report generation plays a key role in assisting medical diagnosis and condition monitoring. The automatic generation of accurate and complete medical imaging reports is of great significance for improving diagnostic efficiency and reducing the burden on doctors. However, traditional rule-based and statistics-based methods are difficult to capture complex language and contextual relationships, and rely heavily on manual intervention, limiting the quality and adaptability of generated reports. To overcome these problems, deep learning-based medical imaging report generation methods have attracted extensive attention in recent years.

The deep learning model for medical report generation usually uses a visual model to encode the spatial and semantic features of medical images, and uses a language model to decode the corresponding paragraph-based document reports, describing important medical findings observed on the corresponding medical images, and emphasizing the size and location of abnormal lesions. Due to the complexity of organ structures and pathological manifestations and the diversity of natural language expressions in medical images, automatic generation of readable and clinically accurate medical image reports is still an unsolved problem. Existing medical report generation methods usually face three major problems:

First, they often use single-view medical images to generate reports~\citep{liu2021exploring,chen2020generating,hou2021ratchet}, but fail to effectively utilize multi-view data. These medical images of different views provide medical information in different directions, and there are rich connections between them~\citep{vu2021medaug}. However, existing methods ignore this inter-view correlation, and the processing of a single view limits the ability of medical report generation models to extract comprehensive information from complex medical data, resulting in reports that may lack accuracy and comprehensiveness. Therefore, some works try to incorporate multi-view medical images~\citep{amjoud2021automatic,xu2020reinforced,xue2018multimodal}, but because the multi-view medical images are simply stitched together as input, there is no in-depth exploration of the relationship between different views, so there is information redundancy and insufficiency to a certain extent.

Furthermore, our experimental study shows that while using multi-view can greatly improve the performance of medical report generation, it suffers from the problem of domain shift. The domain shift problem refers to the situation where the data used in training and the data used in testing/inference have differences in distribution, resulting in poor performance of the model on the test set~\citep{zhang2023multi,ZHANG-MIA2022}. In multi-view medical report generation, when a model uses multi-view data for learning and only single-view data for inference, the domain shift problem occurs, leading to performance degradation.

Moreover, the unimodal optimization method limits the accurate expression of visual information by text description. Medical report generation is usually optimized using a text-based cross-entropy loss, which only focuses on the gap between each predicted word and the real word, and ignores the global coherence and logic of the report. To establish sentence-level matching, \citep{10.1007/978-3-030-32692-0_77,liu2019clinically,li2018hybrid,jing-etal-2019-show,xu2020reinforced} uses reinforcement learning for optimization. Despite further improvements, it is still text-based unimodal matching. These methods cannot realize the semantic alignment of images and texts across modalities, and whether the image content is accurately described is the most important measure of report quality. Therefore, only the multimodal optimization objective can be more suitable for the multimodal task of report generation.

In this paper, to overcome the above problems, we propose a \textbf{C}ross-modal \textbf{C}onsistent \textbf{M}ulti-view medical report generation with \textbf{Do}main \textbf{T}ransfer network (abbreviated by $C^2$M-DoT). Compared to existing medical report generation, the proposed $C^2$M-DoT consists of three improvements.

First, we propose to integrate a Multi-view Contrastive Learning (\textbf{MvCo}) strategy into our previous deep reinforcement learning based medical report generation model~\citep{xu2020reinforced}. Since the paired multi-view medical images are different imaging results of same patient, their descriptions of the lesions or organs in the patient should have high consistency~\citep{vu2021medaug}. Therefore, MvCo is proposed to utilize the semantic embeddings of different views of patients' X-Ray images for contrastive learning. Compared to existing self-supervision based solutions~\citep{yan2021weakly,zhang2020contrastive,azizi2021big} whose contrastive learning modules are applied in encoders, the feature representations used for contrastive learning in MvCo is located in decoders, which uses high-level semantics-based contrastive loss to have a more direct impact on the quality of the final medical report.

Furthermore, to address the domain shift problem due to input distribution, we further propose to incorporate Domain Transfer Network (DoT) into our medical report generation model to address this problem by bridging the performance gap between multi-view and single-view inputs. Specifically, DoT is achieved using a sampling-based adaptive input selection module, which enables the generation branch to randomly select single or multi-view fused features as the final input according to the estimated probability. The advantages of DoT are as follows: (i) it ensures the model is learned using a more comprehensive input distribution (include the single-view and multi-view data input), which thus close the performance gaps of using multi-view and single-view inputs in inference; (ii) Due to the ability to learn based on the characteristics of the input information selects the most appropriate view input at the moment, so using DoT does not reduce the feature learning ability of the model; (iii) Since DoT can probabilistically introduce single-view input into the multi-view generation branch, it also plays a role in narrowing the information gap between the comparative learning branch (with single-view input) and the generation branch(with multi-view input).

Moreover, we introduce cross-modal consistency (CMC) to optimize the model, which extend the original text-based single-modal loss to image-text-based cross-modal loss to obtain medical reports that better match image semantics. Specifically, CMC calculates the semantic similarity of medical images and predicted texts and the semantic similarity of medical images and labeled texts respectively, and further maximizes the consistency of the two semantic similarity matrix distributions. The introduction of multimodal consistency provides a different source of information in addition to the original text report, strengthens the semantic association between images and text, and helps generate reports that more accurately describe image content and improve interpretability.

Overall, the contributions of this paper are as follows:
\begin{itemize}
\item We found that the existing medical report generation methods have problems such as not being able to utilize multi-view data mutual information, domain shift problem, and single-modal optimization, etc., and proposed a cross-modal consistent multi-view medical report generation with domain transfer network ($C^2$M-DoT) to overcome these problems to better generate medical reports .

\item  The improvements in the proposed $C^2$M-DoT are threefold:(i) multi-view contrastive learning (MvCo) strategy, using multi-view information of chest x-ray images for better model learning, (ii) proposing a domain transfer network (DoT), to ensure that the model achieves good performance using only single-view inputs during the inference stage. (iii) Cross-modal optimization (CMC) is introduced to utilize image information to generate medical reports consistent with visual semantics.

\item Extensive experiments are performed on two publicly available medical image reporting benchmark datasets. The experimental results show that our proposed $C^2$M-DoT model greatly outperforms the state-of-the-art baselines in terms of all natural language metrics, and has also achieved remarkable results in image-text matching metrics. 
Secondly, we also conducted ablation experiments to prove that Multi-View Contrast (MvCo), Domain Transfer Network (DoT) and Cross-Modal Optimization (CMC) are effective and necessary conditions for the $C^2$M-DoT model to achieve excellent performance. 
Further experiments show that $C^2$M-DoT can achieve almost the same performance as multi-view input using only single-view input, which can make extensive use of more clinically unpaired incomplete images and avoid excessive exposure of patients to X-rays, saving patients time and money considerably.

\end{itemize}

The remainder of this paper is organized as follows. We present the related works and the details of $C^2$M-DoT in Section~\ref{sec2} and \ref{sec3}, respectively. The experimental studies are introduced in Section~\ref{sec4}, and we summarize this work along with some potential future works in Section~\ref{sec5}.

\section{Related Works}
\label{sec2}

Medical report generation usually uses convolutional neural networks as visual encoders and recurrent neural networks for sentence generation. Due to the emergence of transformers, many works have also used this to improve the quality of long and short text generation~\citep{chen2020generating,hou2021ratchet}. To understand and describe complex lesions more accurately, a series of spatial and linguistic channel attention mechanisms have been carefully designed~\citep{anderson2018bottom, jing-etal-2018-automatic, xue2018multimodal, pan2020x, xu2020reinforced, hou2021ratchet}. At the same time, more additional data and knowledge are often explored to help generate more accurate and comprehensive reports~\citep{liu2021exploring}. However, they all ignore the clinically generated multi-view medical images. In fact, the large associations naturally present in patient metadata benefit the learning of visual representations~\citep{vu2021medaug}. Encouraged by the benefits of multi-view images, \citep{amjoud2021automatic,xu2020reinforced,xue2018multimodal} tried to use multi-view medical images to generate reports, but only concatenated them and directly fed them into the model. This rudimentary method does not deeply explore and utilize the mutual information between different views, which cannot bring great benefits but may cause information redundancy. \citep{yuan2019automatic} further attempts to fuse different views to obtain more information, but due to the generality of the operation, it is not good enough in understanding the semantics of lesions and learning personalized sample features.

Contrastive learning has an excellent performance in learning personalized features of samples. Semantic feature expressions are obtained by comparing anchor samples with positive and negative samples~\citep{chen2020simple,he2020momentum,tian2020contrastive,gao2021simcse}. Usually, augmented forms of the original data are selected as positive samples because they have strong semantic consistency with the original data, such as rotated or cropped images~\citep{chen2020simple}, text sentences replaced by synonyms~\citep{chen2020simple}. Proper selection of positive and negative examples is crucial to the superior performance of contrastive learning. \citep{tian2020contrastive} found that using images from different perspectives of the same natural scene as positive examples can obtain more meaningful visual representations than using augmented images. Different views have greater representational differences but strong consistency at the semantic level.

Therefore, we propose a medical report generation method based on multi-view contrastive learning. Compared with existing methods, the $C^2$M-DoT method proposed in this paper has three advantages: (i) We innovatively use contrastive learning on multi-view medical images, and use contrastive learning for decoded semantic vectors to achieve comprehensive improvement of report quality. (ii) Proposes a domain transfer network that enables multi-view medical report generation models to achieve accurate inference using a single view. (iii) Also adopts cross-modal consistency optimization to enhance the semantic association between reports and images.

High matching between inference reports and original images is an important requirement and ultimate goal of medical report generation. This cross-modal matching mode is commonly found in visual language pre-training models~\citep{radford2021learning}. There is currently a lot of work dedicated to using paired image-text pre-training to improve visual understanding of medical images. \citep{zhang2022contrastive} uses multimodal contrastive learning to change the distance between visual and textual representations in latent space, and then~\citep{yan2021weakly} picks more difficult negative samples on this basis to optimize intra-class difference feature learning. \citep{you2021aligntransformer, huang2021gloria, seibold2022breaking} aligns visual regions and disease labels to learn multi-grained feature representations. \citep{wang2022medclip,wu2023medklip} use medical knowledge-based semantic matching to learn relations between entities. Most of these methods follow Contrastive Language-Image Pre-training (CLIP)~\citep{radford2021learning}, which maximizes the similarity of paired text images while minimizing the similarity of unpaired elements to learn cross-modal matching relations. Furthermore, \citep{eslami-etal-2023-pubmedclip} fine-tunes it on medical data to better adapt CLIP to downstream tasks in the medical field.

Thanks to this efficient multimodal mechanism, we optimize and evaluate the semantic similarity of the inference report to the original image. It is worth noting that: (i) Different from the visual model pre-training work, which separates the upstream visual model training from the downstream medical image analysis tasks, we set the end-to-end learning goal to make the semantic similarity of the image and text of the inference report and The image-text semantic similarity of the real report is consistent, and it is directly optimized for the generation of medical reports. (ii) In cross-modal consistency optimization, we introduce a new supervisory signal: image-text semantic similarity. Compared with the single-modal matching of text based on n-grams, the consistency of semantic similarity between images and texts can capture the subtle semantic differences between texts and promote the semantic consistency between the generated report and the input image. This allows the generated reports to more accurately describe the content and features in the image. (iii) We further explore the combination of this cross-modal consistency with multi-view medical report generation.


\section{Methodology}
\label{sec3}
\begin{figure*}[!t]
\centering
\includegraphics[width=0.9\textwidth]{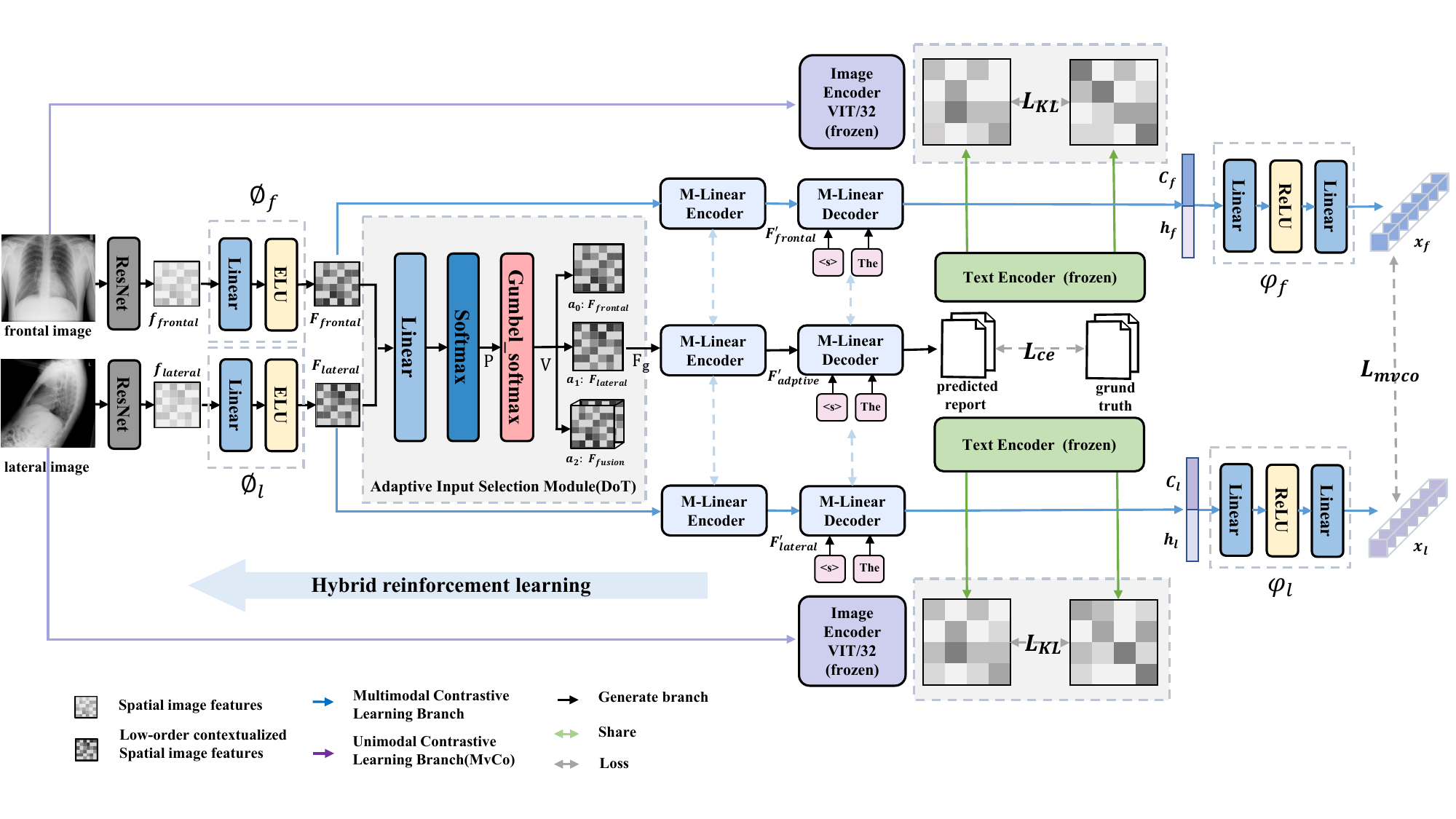}  
\vspace{-1em}
\caption{The architecture of our proposed $C^2$M-DoT network.}
\vspace{-1em}
\label{arch}
\end{figure*}
we propose a Cross-modal Consistent Multi-view medical report generation with Domain Transfer network ($C^2$M-DoT). Intuitively, we believe that the use of multi-view contrastive learning will enhance the model's ability to explore lesion information, resulting in more accurate reports. Besides, we believe that the introduction of the domain transfer network will further bridge the performance gap between different views for report generation, and further improve the comprehensive performance of report generation. Moreover, we believe that combination of the cross-modal consistency optimizations will help generate semantically consistent descriptions and also help improve the correctness of reported results.

Specifically, as shown in Fig.~\ref{arch}, we will adopt the architecture of a two-stream network for multi-view comparative learning and report generation, respectively. The multi-view contrastive learning branch will use frontal and lateral views of medical images independently. In order to focus on the important medical findings of chest X-rays and abstract accurate semantic representations, the original image is first sent to the pre-trained ResNet-101 and Transformer-like encoding network M-Linear encoder, and then the obtained high-order multi-dimensional visual embedding $F^{\prime}$ is sent to the multi-model reasoning network M-Linear decoder. The context information $c$ and hidden state $h$ of the last time step of the decoding process participate in semantic contrastive learning. In the generation branch, medical images from different views will be fed into the domain transfer network together, where the adaptive sampling module will decide the visual embedding $F_g$ that will be finally used to generate report according to the comprehensive information of the current research case. In addition, the decoded predicted report $R_{pre}$ will be further optimized for cross-modal consistency with real report $R_{real}$. We present our network design and implementation details in the following subsections.

\subsection{Multi-View Contrastive Learning}

Although existing medical report generation work has attempted the use of multi-view medical images, no research has explored the positive impact of their extensive associations on medical report generation. Multi-view contrastive learning was first used in natural scenes, and the resulting visual representations achieved excellent performance in downstream tasks such as segmentation and detection. We use multi-view contrastive learning in the task of medical report generation to explore the semantic consistency between different views and help generate reports.

Existing contrastive learning methods are often used in vision model pre-training to optimize representations. However, due to the target differences of upstream and downstream tasks, upstream visual features often cannot generalize well on downstream tasks. To make up for this deficiency, we introduce contrastive learning into an end-to-end medical report generation model to directly compare the decoded semantic embeddings and explore the impact of mutual information between different views of medical images on report generation.

Specifically, we propose a semantic-based multi-view contrastive learning (MvCo) method based on the backbone network of hybrid reinforcement learning medical report generation with M-Linear attention mechanism~\citep{xu2020reinforced}. First, the pre-trained ResNet-101~\citep{he2016deep} is used to initially extract the spatial visual features $f_{frontal}$ and $f_{lateral}$ of the frontal and lateral images of one case. In order to enable multi-view contrastive learning to better utilize the differences of different views for advanced semantically consistent representation learning, we further project the spatial visual features of each view to obtain more distinguishing visual information embeddings.

\vspace{-1em}
\begin{equation} 
F_{frontal}=\phi_f(f_{frontal}) ,F_{lateral}=\phi_l(f_{lateral})
\end{equation}

\noindent where  $\phi_f(\cdot)$ and $\phi_l(\cdot)$ are modeled as fully connected layers with ELU activations. $F_{frontal}$ and $F_{lateral}$ are the frontal and lateral view visual embeddings focusing on the difference of view information, which are fed into two weight-shared report generation networks with m-linear encoder-decoder.

To directly affect the quality of generated reports, multi-view contrastive learning is applied to the decoded semantic embeddings. We concatenate contextual semantic representations $c_f$, $c_l$ and hidden layer information $h_f$, $h_l$ decoded from different views, and then project onto the same implicit space for comparison.

\vspace{-1em}
\begin{equation} 
x_f=\psi(Concat(c_f,h_f)) , x_l=\psi(Concat(c_l,h_l))
\end{equation}

\noindent where $\psi(\cdot)$ is modeled as two fully connected layers with ReLU activations, according to~\citep{nair2010rectified}. We define the similarity between different elements in terms of cosine distance:

\vspace{-1em}
\begin{equation}
\label{cmsim}
    sim(m,n)=\frac{m\cdot n^\top}{\parallel m \parallel \parallel n \parallel},
    sim(n,m)=\frac{n\cdot m^\top}{\parallel n \parallel \parallel m \parallel},
\end{equation}

Since the lesion semantics presented by medical images from different views should be highly consistent, we maximize the similarity between the semantic embeddings of the frontal and lateral views of the same patient, while minimizing the similarity between the semantic embeddings of different patients. The multi-view contrastive loss $L_{MvCo}$ is defined as:

\begin{equation}
\mathcal{L}_{\text{MvCo}} = -\log\frac{\exp(\text{sim}(x_l, x_f)/\tau_c)}{\sum_{k=1}^{2N}\mathbbm{1}_{[k\neq l]}\exp(\text{sim}(x_l, x_k)/\tau_c)}
\end{equation}

\noindent where $\tau_c$ is temperature parameter. In addition, the effects of multi-view contrastive learning using different feature embeddings are detailed in ~\ref{mc}.

\vspace{-1em}
\subsection{Domain Transfer Network}

The excellent properties of multi-view data can improve the semantic expression of the characteristics of abnormal lesions, and help to generate accurate high-quality medical reports. However, an important shortcoming of existing multi-view medical report generation schemes is that since multi-view data complement each other, multi-view data is required not only in the training phase but also in the inference phase, which limits its application in clinical practice. When a multi-view medical report generation model uses single-view data for inference, due to the gap in the input distribution between the training and inference stages, it will inevitably cause performance degradation in report inference, known as the domain shift problem. In order to overcome the above problems, we propose a domain transfer network. During the training process, the model will receive a comprehensive input distribution of various single-view or multi-view images, and adaptively select the current most beneficial input to feed the generative model.

Specifically, we first define the input distribution as an action space $A \in \mathbb{R}^{1\times3}$:

\vspace{-0.5em}
\begin{equation} 
a_i=\left\{
	\begin{aligned}
		F_{frontal} & , & i=0\\
		F_{lateral} & , & i=1\\
            F_{fusion} & , & i=2\\
	\end{aligned}
	\right.
\end{equation}

\noindent where, $F_{frontal}$ and $F_{frontal}$ represent the front and lateral single-view visual feature input respectively, while $F_{fusion}$ represents multi-view view input, which is obtained by adding the features of multi-views instead of concatenating them. The same input width can keep the process consistent between the multi-view generation branch and the contrast learning branch using the front single view respectively to enhance the overall performance of the network.

Then, in order for the model to obtain the most useful information input and better balance the use of frontal and lateral view information, we adaptively decide to input a single feature or a mixture of features through action sampling. This form of non-determinism enables the model to adaptively select the best input to obtain the maximum amount of visual information for each image.


\begin{figure*}[!t]
\centering
\includegraphics[width=0.95\textwidth]{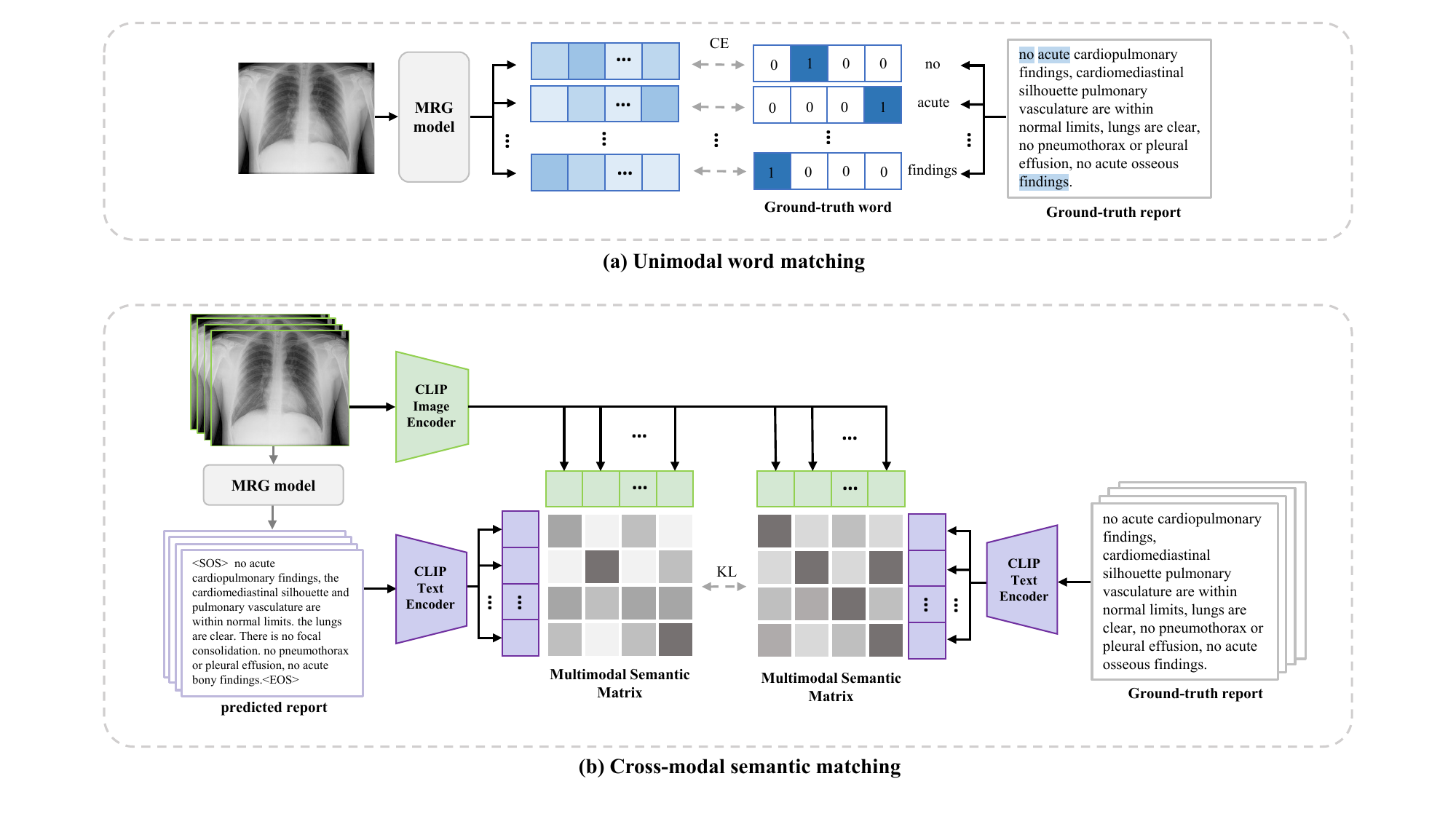}
\vspace{-2em}
\caption{Schematic diagram of (a) traditional single-modal optimization based on word matching and (b) our cross-modal optimization based on consistency of image-to-text semantic similarity matrix.}
\label{cont}
\end{figure*}
To circumvent the technical problem that binary sampling actions cann't be differentiated to participate in backpropagation, we utilize random sampling based on \emph{Gumbel-Softmax} distribution. This reparameterization trick has been used in reinforcement learning for making discrete decision ~\citep{jang2016categorical}. Non-differentiable action values will be replaced by differentiable samples from the \emph{Gumbel-Softmax} distribution. 
Specifically, we concatenate the global visual features $F_{frontal}$ and $F_{lateral}$, which are multi-scale fusions of frontal and lateral views in the latent space and taken as a comprehensive information basis for the current action selection. It is sent to a linear layer through the fully connected layer to obtain the action confidence warehouse $P \in \mathbb{R}^{1\times3}$.

\begin{equation} 
P=softmax(W_c(Concat(F_{frontal},F_{lateral})))
\end{equation}

\noindent where $W_c$ represents the fully connected layer parameter matrix. Subsequently, the sampling module will generate action values $V \in \mathbb{R}^{1\times3}$, which defined as

\begin{equation} 
\hspace{-0.5em}
V(a)=\frac {exp((\log(P_i(a)+g_i(a)/\tau_s)} {\sum_{j=1}^{3}{exp((\log(P_j(a)+g_j(a)/\tau_s)}},for\ i=1,2,3
\end{equation}

\noindent where $g$ represents the noise sampled from the standard \emph{Gumbel-Softmax} distribution, and $\tau_s$ is the temperature parameter. The final input strategy is gained after $V$ through \emph{argmax} layer. During the inference stage, $V$ is generated according to the input directly. Sample the action whose sample value in $A$ is calculated to be $1$, and reconstruct only the features corresponding to the action into the final input feature $F_g$.

\begin{equation} 
F_g= A(a_i ), V(a_i )=1,
\end{equation}

\subsection{Cross-modal consistency}

Traditional medical report generation methods have always used cross-entropy as an optimization method. As shown in Fig.~\ref{cont} (a), assuming the real report text sequence $R=\left\{ r_1, r_2,..., r_Q \right\}$, and the probability distribution sequence of the text sequence generated by the model $B=\left\{ b_1, b_2,..., b_Q \right\}$. The cross-entropy loss is calculated as:

\begin{equation} 
L_{CE}(R,B) = -\sum_{q=1}^{Q} \sum_{d=1}^{D} r_{q,d} \cdot \log(b_{q,d})
\end{equation}

\noindent where $D$ is the size of the vocabulary, and $Q$ is the sequence length. $b_{q,d}$ represents the probability that the word at the $q^{th}$ position in the sequence is the $d^{th}$ word in the vocabulary, while, $r_{q,d}$ is a one-hot vector, indicating whether the word at the $q^{th}$ position in the real sequence is the $d^{th}$ word in the vocabulary. Only one element of this vector is $1$, which represents the real word position, and the other elements are $0$.

The cross-entropy loss improves the quality of generated text by minimizing the difference between the probability distribution generated by the model and the real text distribution. However, since the calculation of each time step of the text sequence is independent, it can only focus on a single word, but cannot effectively capture the contextual relationship between words; moreover, all words in the text are mapped to numerical values representing probabilities, completely ignoring the semantic information of the text.

Although there are many reinforcement learning methods that use natural language indicators as rewards to further optimize the model and better capture the relationship between text contexts to improve coherence, only relying on limited language pattern matching still cannot truly understand the semantics of text.
In addition, previous optimization methods are usually limited to the processing of single-modal text data, which means that these methods mainly focus on text information and ignore other important medical data modalities (such as medical images), and cannot effectively adapt to the multi-modal reasoning task of medical report generation. Therefore, we introduce cross-modal consistency loss to enhance the semantic consistency relationship between reports and images to ensure that the generated reports correctly describe and interpret image-related information.

Specifically, we define a visual encoder $V_{CLIP}$ and a language encoder $T_{CLIP}$ to extract features of medical images and report texts respectively.

Referring to the setting of PubMedCLIP~\citep{eslami-etal-2023-pubmedclip}, we use the VIT-B/32 Vision Transformer~\citep{dosovitskiy2020image} fine-tuned by medical dataset Radiology Objects in COntext (ROCO)~\citep{pelka2018radiology} to encode the frontal and lateral original images into visual features $v_f$and $v_l$; and encode the generated reports and real reports into textual features$t_{pre}$ and $t_{real}$:

\vspace{-0.5em}
\begin{equation}
v_f = V_{CLIP}(I_f) , v_l = V_{CLIP}(I_l),
\end{equation}
\begin{equation}
t_{pre} = T_{CLIP}(R_{pre}) , t_{true} = T_{CLIP}(R_{true}),
\end{equation}

\noindent where, $I_f$ is the frontal image, $I_l$ is the lateral image, $R_{pre}$ is the prediction report, and $R_{true}$ is the real report. Then, we define the similarity between two modalities via the cosine phase similarity distance shown in Equation~\ref{cmsim} and apply softmax normalization to it. 

\begin{equation} 
SFP^{v2t}_{i}=\frac{exp(sim(v_f,{t_{pred}}_i)/\tau_m)}{\sum_{j=1}^{N}{\ {exp(sim(v_f,{t_{pred}}_j)/\tau_m)}}},\\
\end{equation}
\begin{equation} 
SFP^{t2v}_{i}=\frac{exp(sim(t_{pred},{v_f}_i)/\tau_m)}{\sum_{j=1}^{N}{\ {exp(sim(sim(t_{pred},{v_f}_j)/\tau_m)}}},\\
\end{equation}
\begin{equation} 
SFT^{v2t}_{i}=\frac{exp(sim(v_f,{t_{true}}_i)/\tau_m)}{\sum_{j=1}^{N}{\ {exp(sim(v_f,{t_{true}}_j)/\tau_m)}}},\\
\end{equation}
\begin{equation} 
SFT^{t2v}_{i}=\frac{exp(sim(t_{true},{v_f}_i)/\tau_m)}{\sum_{j=1}^{N}{\ {exp(sim(sim(t_{true},{v_f}_j)/\tau_m)}}},\\
\end{equation}

\noindent where $\tau_m$ is a learnable temperature parameter, and $N$ is the number of training pairs. $SFP^{v2t}_{i}$, $SFP^{t2v}_{i}$, $SFT^{v2t}_{i}$ and $SFT^{t2v}_{i}$ are softmax normalized similarity scores from frontal image to predicted text, predicted text to frontal image, frontal image to real text, and real text to frontal image. Similarly, the normalized similarity scores from lateral image to predicted text, predicted text to lateral image, lateral image to real text and real text to lateral image are calculated as $SLP^{v2t}_{i}$, $ SLP^{t2v}_{i}$, $SLT^{v2t}_{i}$ and $SLT^{t2v}_{i}$, respectively.


Since the major human organs and obvious abnormalities in the X-rays are described in the report, there are inevitably more or less semantic similarities. Therefore, it is unreasonable to simply maximize the diagonal similarity of the similarity matrix to 1 and minimize the off-diagonal similarity to 0. Finally, we use Kullback–Leibler (KL) to optimize the similarity matrix. Our goal is to make the similarity matrix between the predicted text and the image close to the similarity matrix between the real text and the image:

\vspace{-0.5em}
\begin{equation} 
\mathcal L_{CMC}^{F}=\frac{1}{2}{\epsilon_{{(v,t)}\sim\Theta}(KL(SFP^{v2t}_{i},SFT^{v2t}_{i})+KL(SFP^{t2v}_{i},SFT^{t2v}_{i}))}
\end{equation}
\noindent where $L_{CMC}^{F}$ is the frontal image text similarity loss, and the lateral image text similarity loss $L_{CMC}^{L}$ is calculated in a similar way. Finally, similarity losses for image and text modalities are combined with multi-views for cross-modal consistency optimization.

\vspace{-0.5em}
\begin{equation} 
\mathcal L_{CMC} = L_{CMC}^{F} + L_{CMC}^{L}.
\end{equation}

\section{Experiments}
\label{sec4}

\subsection{Datasets}

To evaluate the performance of our proposed $C^2$M-DoT, extensive experiments are conducted on two publicly available datasets. As shown in Table~\ref{dataset}, both datasets contain chest X-ray images and paired free-text reports.

(i) IU X-Ray~\citep{demner2016preparing} is one of the most commonly used medical image description datasets, collected by Indiana University. (ii) MIMIC-CXR~\citep{johnson2019mimic} is currently the largest publicly available medical image description dataset, proposed by the Massachusetts Institute of Technology. Each imaging study may contain one or more images, including posteroanterior (PA) or anteroposterior (AP) views and lateral (LL) views. The medical report corresponding to the imaging results consists of multiple sentences, in which the two parts \textit{impression} and \textit{findings} summarize the main diagnostic results.

We preprocess the above two datasets as follows: First, for multi-view contrastive learning, we filter the data to only retain cases with frontal and lateral medical images and complete reports. The number of cases in each dataset is: (i) IU X-Ray: 6,222 images, 3,111 corresponding reports, (ii) MIMIC-CXR: 153,448 images, 76,724 corresponding reports. Then, we resize the image to 224x224. For reports, \textit{impression} and \textit{findings} will be generated simultaneously. We convert all words to lowercase and remove special characters. Thereafter, we tokenize the reports to build word lists. In order to filter out many uncommon words, simplify the model structure and prevent overfitting, we only keep the words that appear more than 5 times, and replace the discarded words with the UNK token. The number of word tokens per dataset is: (i) IU X-Ray: 776 tokens, (ii) MIMIC-CXR: 2991 tokens.

Finally, for all datasets, randomly select $70\%$ of the datasets for training, $10\%$ for validation, and $20\%$ for testing, and make sure there is no overlap between datasets.

\vspace{-1em}
\begin{table}[h] 
  \caption{Datasets Information.}
  \vspace{0.5em}
      \centering
  \begin{tabular}{cccc}
   \toprule
   Datasets & Images & Views & Reprots  \\
   \midrule
   IU X-Ray & 7,470 & Multi & 3,955 \\
   MIMIC-CXR & 377,110 & Multi & 227,827  \\
   \bottomrule
   \end{tabular}
   \label{dataset}
\end{table}
\vspace{-1em}

\subsection{Evaluation Metrics}
In order to evaluate the quality of the report, we use the six most commonly used metrics for automatic language generation for comprehensive evaluation, including BLEU-n~\citep{papineni2002bleu} , METEOR~\citep{banerjee2005meteor} and ROUGE-L~\citep{lin2004rouge}, where BLEU-n refers to four n-gram-based indicators (BLEU-1 to BLEU-4). Specifically, BLEU (Bilingual Evaluation Study) is used to measure exact matches, which counts the number of n-grams that match between the generated report and the reference standard, computing a weighted score. METEOR (Metric for Evaluation of Translation with Explicit ORdering) is a comprehensive evaluation metric that further considers word order information in addition to accuracy. Rouge-L is a Longest Common Subsequence (LCS)-based metric that combines the length of the longest common subsequence with the length of the reference text to produce a score for measuring the similarity between the generated report and the reference standard.


\subsection{Baselines}

We compare our method with six state-of-the-art image captioning and medical report generation models: 
(i) our re-implementation of the top-down model \citep{anderson2018bottom}, which is a classic encoder-decoder-based model for image captioning employing a conventional attention mechanism that calculates the contribution of regional features to the texts to be generated, and  
(ii) MRMA~\citep{xue2018multimodal}, an encoder-decoder-based model specially designed for medical report generation, in which reports are generated sentence by sentence with a recurrent way to generate long paragraphs. 
(iii) RTMIC~\citep{10.1007/978-3-030-32692-0_77}, which is a state-of-the-art medical report generation method based on reinforcement learning, enhancing the capacity of the generation model with reinforcement learning, and increasing the clinical accuracy with a transformer. 
(iv) X-LAN~\citep{pan2020x}, which is an image captioning model employing x-linear attention and improving it with reinforcement learning. As image captioning is similar to our task to some extent, we also take this model as our baseline. 
(v) HReMRG-MR~\citep{xu2020reinforced}  which is a medical report generation model that utilizes a hybrid reinforcement learning method and uses a high-order attention mechanism to repeat the penalty mechanism to improve reports.
(vi)R2Gen~\citep{chen2020generating}, a memory unit-based medical report generation method. Models and memorizes similar patterns between reports, thereby facilitating Transformer to generate more informative long-text explanation reports.
For our implemented methods, we use the same visual features and train/val/test split on both datasets.

\subsection{Implementation Details}

We utilize ResNet-101 pre-trained on ImageNet~\citep{deng2009imagenet} to extract $2048$ dimensional region-level image features from the last convolutional layer. After being converted to visual embeddings of size $1024$, the encoder exploration with four stacks of M-linear attention blocks yields high-order synthetic features. During the decoding process, we set the size of hidden layer, word embedding dimension, and the latent dimension of the projection layer to $1024$. During training, we first pre-train the model with a batch size of $6$ for $60$ epochs using NVIDIA RTX 2080Ti GPUs. 
We set the base learning rate to $0.0001$, paired with a Norm decay strategy with $10,000$ warm-up steps, and used the ADAM~\citep{kingma2014adam} optimizer. We set $\tau_c$ to $0.1$ and $\tau_s$ to $0.3$. Finally, we train the model with the batch size of $2$ for $60$ epochs of reinforcement learning~\citep{rennie2017self} using beam search~\citep{vijayakumar2016diverse} with a beam size of $2$ to further improve the model performance. We set the indicator-weighted mixed reward as our training reward~\citep{xu2020reinforced}, where the weights of BLEU-1, BLEU-2, BLEU-3, BLEU-4, METERO, and ROUGE-L, are $2$, $2$, $1$, $1$, $2$, and $2$, respectively; and the base learning rate is reduced to $0.00001$ and decayed by cosine annealing with a period of $15$ epochs.

\subsection{Main Results}

\begin{table*}[!t] 
	\centering 
 
    \caption{Experimental results of $C^2$M-DoT and the state-of-the-art baselines on IU X-Ray (upper part) and MIMIC-CXR (lower part).}
    \vspace*{1em}
    \resizebox{0.85\textwidth}{!}{
    \begin{tabular}{c|c|cccccc}
		\hline  
  
		Dataset & Model & BLEU-1 & BLEU-2 & BLEU-3 & BLEU-4   & METEOR & ROUGE-L    \\ 
  
             \hline 
		\multirow{6}{*}{\scriptsize{\textbf{IU X-Ray}}} 
		& Top-down & 0.2822 & 0.1866 & 0.1241 & 0.0830 & 0.1455 & 0.3330   \\ 
		& MRMA & 0.3820 & 0.2520 & 0.1730 & 0.1200 & 0.1630 & 0.3090  \\ 
		& RTMIC  & 0.3448 & 0.2188 & 0.1484 &0.1063  &0.1630 &0.3090 \\
		& X-LAN  & 0.3826 & 0.2724 & 0.1949 & 0.1405 &0.1750 & 0.3441   \\
		& HReMRG-MR & 0.4265 & \underline{0.3025} & \underline{0.2119} & 0.1502 & \underline{0.1871} & \underline{0.3608}   \\
		& R2Gen & \underline{0.4349} & 0.2802 & 0.1868 & \underline{0.1510} &  0.1773 & 0.3509   \\
		& $C^2$M-DoT(ours) & \textbf{0.4579} & \textbf{0.3214} & \textbf{0.2302} & \textbf{0.1593} &  \textbf{0.2037} & \textbf{0.3803}  \\	
		\hline
		\multirow{6}{*}{\scriptsize{\textbf{MIMIC-CXR}}}  
		&Top-down & 0.2371 & 0.1548 & 0.1201 & 0.0989 &  0.1352 & 0.3211 \\ 
		&MRMA  &0.3610 & 0.2440 & 0.1820 & 0.1410 &  0.1570 & 0.3300    \\
		&RTMIC &0.3701 &0.2490 &0.1812 &0.1299  &0.1506 &0.3276  \\
		&X-LAN &0.3656 &0.2670 &0.1881 &0.1315  &0.1703 &0.3421 \\
		& HReMRG-MR & 0.4696 & \underline{0.3251} & \underline{0.2412} & 0.1877  & \underline{0.1993} & \underline{0.3742} \\
		& R2Gen & \underline{0.4700} & 0.3098 & 0.2390 & \underline{0.1911}  & 0.1905 &  0.3609   \\
		& $C^2$M-DoT(ours) & \textbf{0.4842} & \textbf{0.3450} & \textbf{0.2579} & \textbf{0.1925}  & \textbf{0.2098} & \textbf{0.3850}  \\
		\hline 
        \end{tabular} }
	\label{Table:baseline} 
\end{table*}

\begin{figure*}[!t]
\centering
\includegraphics[width=0.85\textwidth]{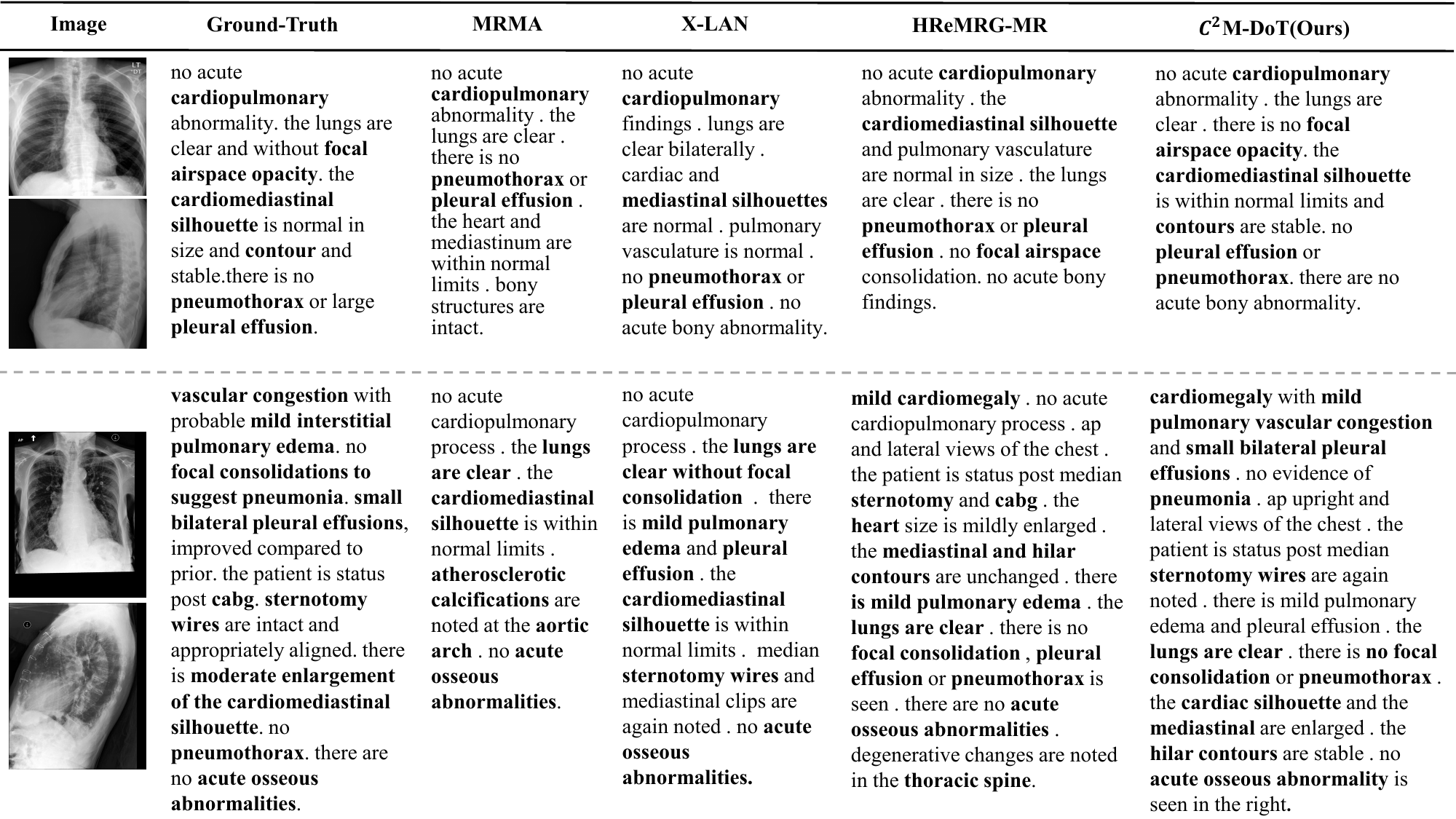} 
    \vspace*{-1em}
\caption{Example of reports generated by our $C^2$-DoT model and baselines.
} 
    \vspace*{-1em}
\label{fig:baseline}    
\end{figure*}

Table~\ref{Table:baseline} shows the experimental results of our proposed $C^2$M-DoT and six baselines on six natural language generation metrics, where all baselines are re-implemented by us. Furthermore, Fig.~\ref{fig:baseline} presents some examples of reports generated by these models.

In general, $C^2$M-DoT outperforms all state-of-the-art baselines among all natural language evaluation metrics in Table \ref{Table:baseline}, and Fig.~\ref{fig:baseline} shows that $C^2$M-DoT also generates more comprehensive and accurate reports (with more matches).

Specifically, in Table~\ref{Table:baseline}, Top-down and MRMA perform poorly in long text generation without reinforcement learning; as shown in Fig.~\ref{fig:baseline}, RTMIC and X-LAN cannot use high-order attention modules to capture visual features for multimodal reasoning, and the report accuracy is low; R2Gen achieves the highest BLEU score among all baselines due to the use of memory units to generate coherent reports; HReMRG-MR uses hybrid reinforcement learning and generally improves on most metrics, and is the best performer on METEOR and ROUGE-L among all baselines. On this basis, $C^2$M-DOT achieves the best results on all metrics because (i) our multi-view contrastive learning adequately performs multi-view mutual information learning to obtain superior performance, (ii) the same input distribution for multi-view training and single-view testing is maintained, and the task gap between contrastive learning and generation branches is narrowed, avoiding the domain shift problem. (iii) Cross-modal consistency optimization makes inference report semantics and image semantics consistent.


\subsection{Ablation Study}

\begin{table*}[!t] 
	\centering  
\caption{Automatic natural language evaluation on IU X-Ray (upper part) and MIMIC-CXR (lower part). Ablation studies, where MvCo indicates multi-View contrastive learning, DoT indicates domain transfer network, and CMC indicates Cross-modal consistency. The best results are highlighted. }
 \vspace*{1em}
  \resizebox{0.85\textwidth}{!}{
	\begin{tabular}{c|c|cccccc}
		\hline  
		Dataset & Model & BLEU-1 & BLEU-2 & BLEU-3 & BLEU-4   & METEOR & ROUGE-L     \\ \hline 
		\multirow{6}{*}{\scriptsize{\textbf{IU X-Ray}}} 
		& Base-Cat & 0.4175 & 0.2813 & 0.1915 & 0.1400 &0.1820 & 0.3604  \\
            & MvCo-Cat & 0.4373 & 0.3062 & 0.2139 & 0.1482  & 0.1933 & 0.3609  \\
		& MvCo-Fus & 0.4440 & 0.3130 & 0.2196 & 0.1571  & 0.1953 & 0.3698   \\
		& MvCo-DoT & 0.4533 & 0.3180 & 0.2228 & 0.1568  & \underline{0.1958} & 0.3743   \\
            & MvCo-CMC & \textbf{0.4581} & \underline{0.3193} & \underline{0.2245} & \underline{0.1583} & 0.1934 & \underline{0.3772}   \\
            & $C^2$M-DoT (Ours) & \underline{0.4579} & \textbf{0.3214} & \textbf{0.2302} & \textbf{0.1593} & \textbf{0.2037} & \textbf{0.3803}  \\
		\hline
		\multirow{6}{*}{\scriptsize{\textbf{MIMIC-CXR}}}  
		& Base-Cat & 0.4380 & 0.2995 & 0.2132 & 0.1626  & 0.1817 & 0.3647  \\
            & MvCo-Cat & 0.4521 & 0.3153 & 0.2389 & 0.1704 & 0.1896 & 0.3670  \\
            & MvCo-Fus & 0.4671 & 0.3211 & 0.2362 & 0.1668  & 0.1935 & 0.3697  \\
            & MvCo-DoT & 0.4698 & \underline{0.3286} & 0.2416 & 0.1792 & \underline{0.1984} & 0.3823  \\
            & MvCo-CMC & \underline{0.4772} & 0.3268 & \underline{0.2423} & \underline{0.1856}  & 0.1908 & \textbf{0.3859}  \\
            & $C^2$M-DoT (Ours) & \textbf{0.4842} & \textbf{0.3450} & \textbf{0.2579} &\textbf{0.1925}  & \textbf{0.2098} & \underline{0.3850} \\
		\hline 
	\end{tabular} }
	\label{Table:abalationStudy} 
\end{table*}
In this section, we report on a series of ablation experiments, using $C^2$M-DoT and five incrementally implemented intermediate models to show the effectiveness of using the proposed multi-view contrastive learning, domain transfer module and cross-modal consistency in our work. Specifically, we implement five incrementally implemented intermediate models: (i) We take the reinforcement learning-based report generation model as the base model, and use the concatenated features of different views as the input, called Base-Cat; (ii) introduce a multi-view contrastive learning branch on the base model, called MvCo-Cat; (iii) A multi-view contrastive learning model using fused features from different views as input, called MvCo-Fus; (iv) Using a domain transfer network capable of adaptively selecting inputs based on multi-view contrastive learning, called MvCo-DoT; (v) introducing a cross-modal consistency loss based on multi-view contrastive learning, called MvCo-CMC. In the Table \ref{Table:abalationStudy} we compare the results of the above six models. 

\subsubsection{Effectiveness of Multi-View Contrastive Learning}

By comparing the results of Baes-Cat and MvCo-Cat, we find that using contrastive learning on multi-view medical images in the multi-view base model makes MvCo-Cat significantly outperform Baes-Cat on all natural language metrics. This finding demonstrates that the mutual information mined by the proposed multi-view contrastive learning helps focus on salient lesions, explore deep semantic features and enable multimodal reasoning. In addition, MvCo-Fus has further improved the results compared to MvCo-Cat, which also shows that the fusion of visual features of different views as the input of the generative model is more suitable for multi-view generation tasks than direct Concatenating. Concatenating different views will double the width of multi-view input features, while fusion features can keep the same width as single-view features. MvCo-Fus narrows the input distribution difference between the contrastive learning branch for single-view input and the generation branch for multi-view input, thus achieving better performance.
 
\begin{figure}[!t]
\vspace{0.5em}
\begin{minipage}[a]{.48\linewidth}
  \centering
  \centerline{\includegraphics[width=4.0cm]{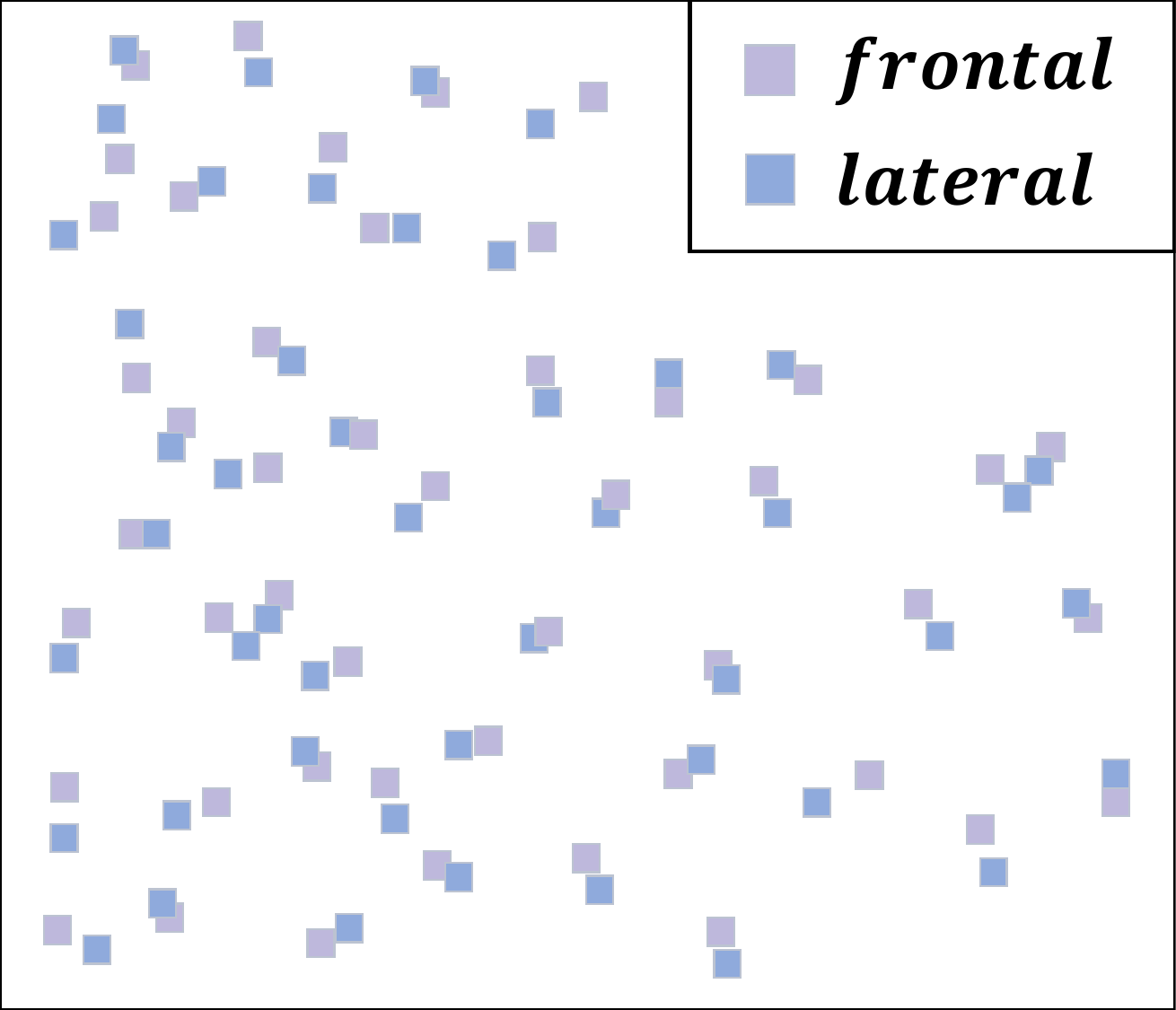}}
  \centerline{(a) Base-Cat}\medskip
\end{minipage}
\hfill
\begin{minipage}[a]{0.48\linewidth}
  \centering
  \centerline{\includegraphics[width=4.0cm]{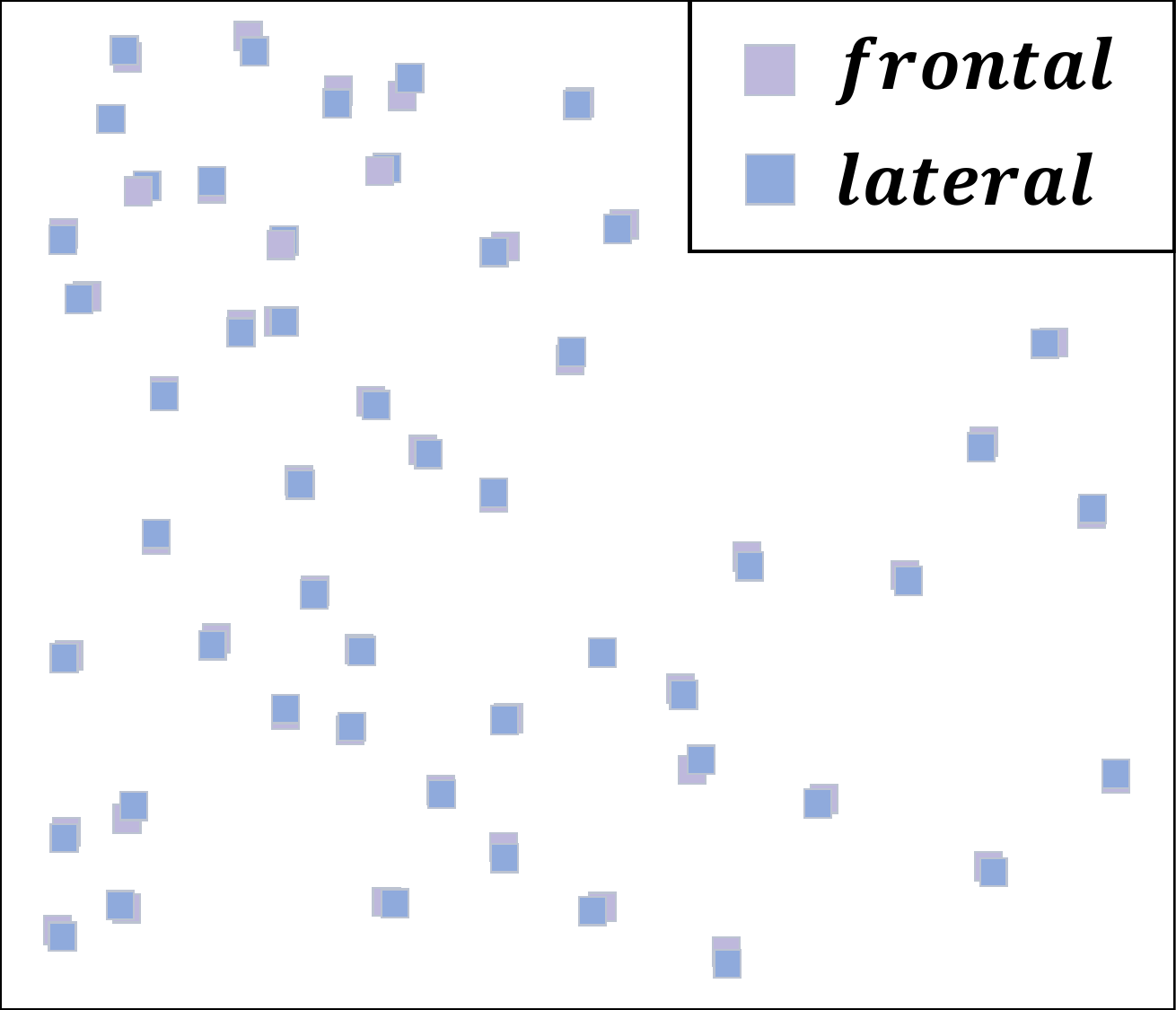}}
  \centerline{(b) MvCo-Fus}\medskip
\end{minipage}
\vspace{-1em}
\caption{Comparison of semantic feature embeddings between Base-C and MvCo-F in latent space.
}
\vspace{-1em}
\label{MvCo}
\end{figure}

Additionally, the effectiveness of multi-view contrastive learning can be visualized in Fig.~\ref{MvCo}, which shows the semantic embeddings for paired multi-view medical image (front, side) decoding. We randomly select 50 pairs of images from the IU-X-Ray dataset. Through the multi-view comparison learning model, the report semantics generated by the medical image is decoded, and mapped to the same latent space to obtain the feature embedding. 
Then, t-SNE is used to reduce the dimensionality of the features in these high-dimensional spaces to represent them in 2D images. The closer the frontal semantic (purple) and corresponding lateral semantic (blue) embeddings are, the better the learned multi-view consistency features are. We observe that frontal and side-reported semantic features for the same patient are closer in the latent space of MvCo-Fus compared to Base-Cat. Therefore, the use of multi-view contrastive learning effectively enhances the learning of mutual information between multiple views to decode consistent semantic feature embeddings.

\begin{figure}[!t]
\begin{minipage}[a]{.48\linewidth}
  \centering
  \centerline{\includegraphics[width=4.2cm]{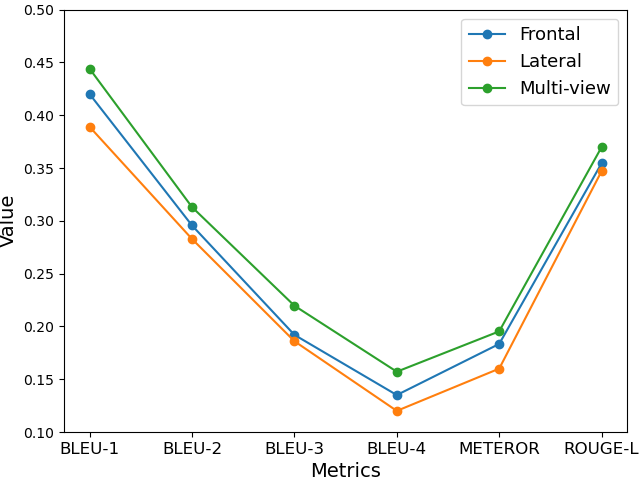}}
  \centerline{(a) MvCo-Fus}\medskip
\end{minipage}
\hfill
\begin{minipage}[a]{0.48\linewidth}
  \centering
  \centerline{\includegraphics[width=4.2cm]{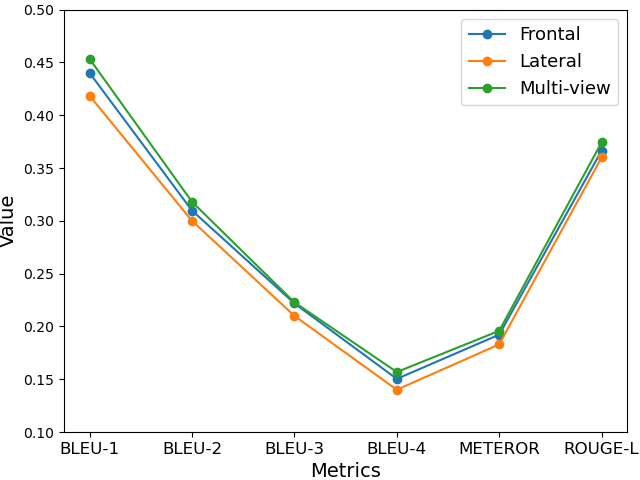}}
  \centerline{(b) MvCo-DoT}\medskip
\end{minipage}

\begin{minipage}[a]{0.48\linewidth}
  \centering
  \centerline{\includegraphics[width=4.2cm]{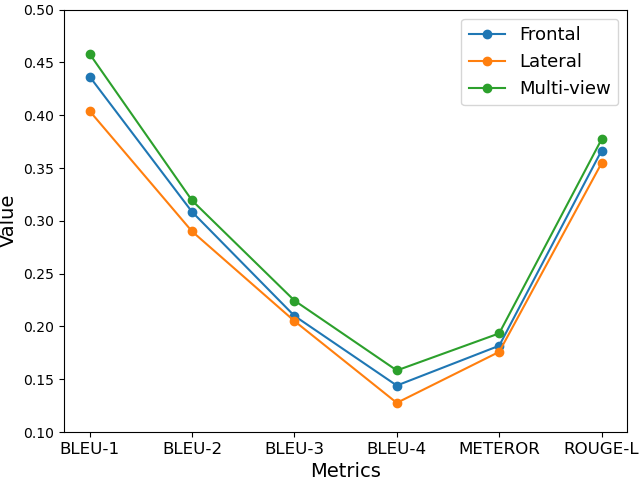}}
  \centerline{(c) MvCo-CMC}\medskip
\end{minipage}
\hfill
\begin{minipage}[a]{0.48\linewidth}
  \centering
  \centerline{\includegraphics[width=4.2cm]{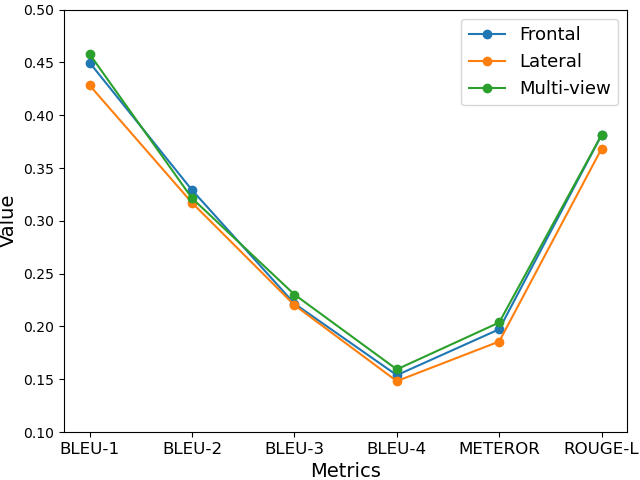}}
  \centerline{(d) $C^2$M-DoT}\medskip
\end{minipage}

\vspace{-1em}
\caption{Performance comparison of models using various view inputs.}
\label{DoT}
\end{figure}
\begin{figure}[!t]
    \centering
    \includegraphics[width=0.48\textwidth]{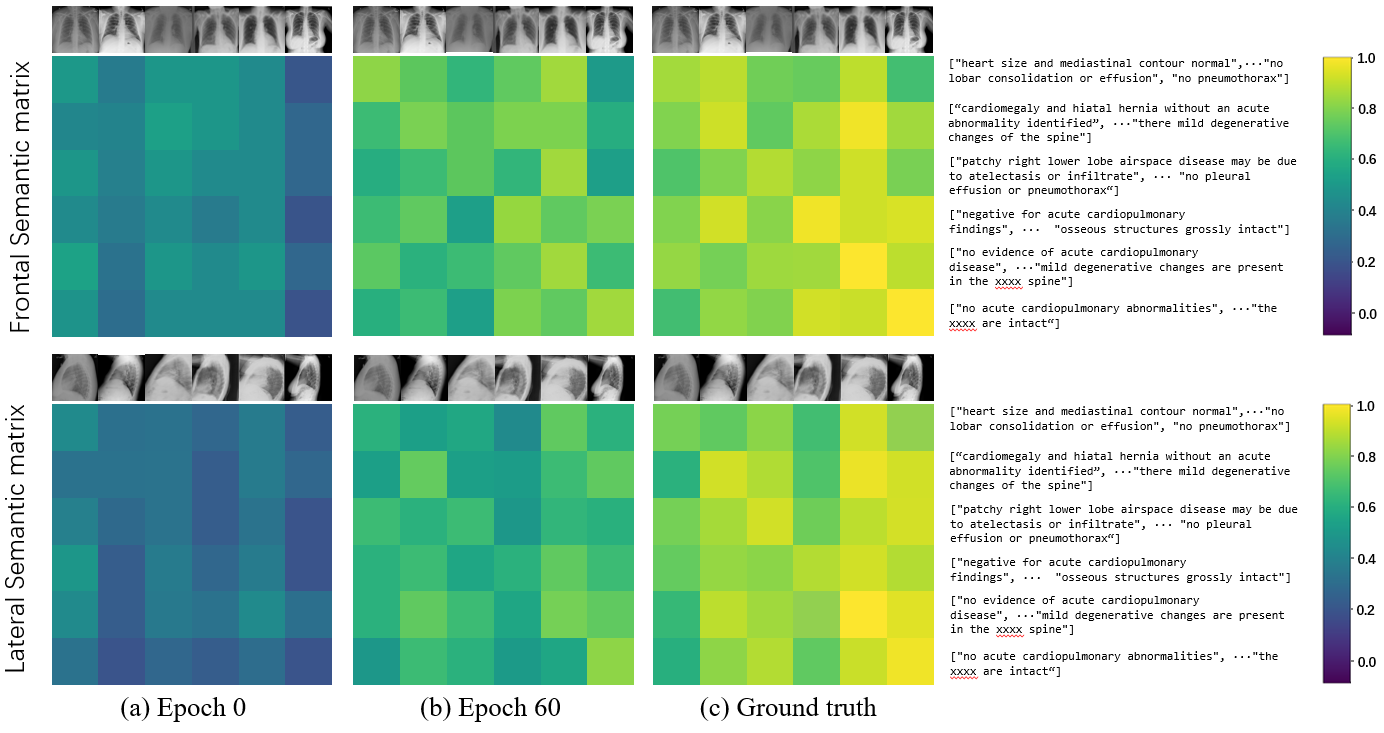} 
    \vspace{-2em}
    \caption{Variation of Multimodal Semantic Similarity Matrix During Training Phase.
    } 
    \label{fig:multisim}   
\end{figure}

\subsubsection{Effectiveness of domain transfers}

Next, we use MvCo-Fus and MvCo-CMC as baselines to compare with $C^2$M-DoT and MvCo-DoT, respectively, to demonstrate the effectiveness of domain transfer networks. In the Table~\ref{Table:abalationStudy}, the results show that whether it is the single-modal optimization model MvCo-DoT or the multi-modal optimization model $C^2$M-DoT, the results after using the domain transfer network are better than the baselines. The adaptive input selection module selects the most suitable input for the model according to the actual situation to maximize the learning benefits; at the same time, since the generative model is provided with the option of single-view feature input during the training process, the multi-view generation process will turn to single-view reasoning under a certain probability. The more identical input distribution narrows the task gap between the generative branch and the multi-view contrastive learning branch (single-view inference), and the model obtains the optimal representation of both, enabling the two processes to promote each other.

Furthermore, the effectiveness of the domain transfer module can be visualized in Fig.~\ref{DoT}, which shows the performance comparison of MvCo-F and MvCo-CMC when using single-view and multi-view inputs before (eg.(a)(c)) and after (eg.(b)(d)) introducing the domain transfer network. Our further observation: Compared with MvCo-Fus and MvCo-CMC, C2M-DoT and MvCo-DoT introduce a domain transfer network for adaptive input selection during training, which solves the problem of domain shift caused by different input distributions during multi-view training and single-view testing. The models can improve cross-domain transferability and can generate high-score reports given any view as input.

\subsubsection{Effectiveness of Cross-Modal Consistency}
\begin{table*}[!t] 
\caption{Comparison of different input sampling methods for domain transfer networks.}
	\centering   
     \resizebox{0.95\textwidth}{!}{
	\begin{tabular}{c|c|cc|cccccc}
		\hline 

   \multirow{2}{*}{Dataset} & \multirow{2}{*}{Model} & \multicolumn{2}{c|}{Input}  & \multirow{2}{*}{BLEU-1} & \multirow{2}{*}{BLEU-2}& \multirow{2}{*}{BLEU-3}& \multirow{2}{*}{BLEU-4}& \multirow{2}{*}{METEOR}& \multirow{2}{*}{ROUGE-L}\\ \cline{3-4}
   & & Frontal&Lateral& &  & &  &  &   \\\hline 
	\multirow{9}{*}{\scriptsize{IU X-Ray}}
	   &\multirow{3}{*}{\scriptsize{DoT-Argmax}} 
		&\checkmark & \checkmark & 0.4647& 0.3296& 0.2317& 0.1603& 0.1926& 0.3722 \\
            & &\checkmark &   & 0.4404 & 0.2992 & 0.2176& 0.1449  & 0.1862 & 0.3636 \\
            & & & \checkmark  & 0.4283 & 0.3171 & 0.2202& 0.1482  & 0.1857 & 0.3680 \\
            \cline{2-10}
          &\multirow{3}{*}{\scriptsize{DoT-Random}}
		  &\checkmark & \checkmark & 0.4385 & 0.2845 & 0.1751& 0.1239&0.1811&0.3124 \\
             & &\checkmark &  & 0.4382 & 0.2859 & 0.2095& 0.1392&0.1688&0.3259 \\
             & & & \checkmark & 0.4108 & 0.2762 & 0.1743& 0.1138&0.1631&0.3225 \\
             \cline{2-10}
          &\multirow{3}{*}{\scriptsize{DoT-Gumbel}} 
		&\checkmark & \checkmark & 0.4579& 0.3214& 0.2302& 0.1593&  0.2037& 0.3803 \\
            & &\checkmark & & 0.4498 & 0.3291 & 0.2216 & 0.1538  & 0.1973 & 0.3813 \\
            & & & \checkmark  & 0.4679 & 0.3291 & 0.2302 & 0.1593  & 0.2037 & 0.3813 \\
		\hline
       \multirow{9}{*}{\scriptsize{MIMIC-CXR}}
         &\multirow{3}{*}{\scriptsize{DoT-Argmax}}
		 &\checkmark & \checkmark & 0.4874 & 0.3490 & 0.2611& 0.1984& 0.1915 & 0.3640  \\
          & &\checkmark &  & 0.4748 & 0.3189 & 0.2372& 0.1726 & 0.1906 & 0.3576 \\
          & & & \checkmark & 0.3934 & 0.2892 & 0.1777& 0.1244  & 0.1607 & 0.3005 \\
          \cline{2-10}
         &\multirow{3}{*}{\scriptsize{DoT-Random}}
	   &\checkmark & \checkmark & 0.4629 & 0.32075 & 0.2055& 0.1639&0.1801&0.3324 \\
          & &\checkmark &  & 0.4637 & 0.3153 & 0.2268& 0.1769&0.1702&0.3222 \\
          & & & \checkmark & 0.4491 & 0.2974 & 0.2050& 0.1437&0.1681&0.3109 \\
          \cline{2-10}
         &\multirow{3}{*}{\scriptsize{DoT-Gumbel}}
	   &\checkmark & \checkmark & 0.4842& 0.3450& 0.2579& 0.1925& 0.2098& 0.3850\\
	   & &\checkmark &  & 0.4712 & 0.3446 & 0.2518 & 0.1916  & 0.2048 & 0.3871 \\ 
          & & & \checkmark  & 0.4520 & 0.3305 & 0.2401 & 0.1907  & 0.1996 & 0.3624  \\
	   \hline 
	\end{tabular}} 
	\label{Table:dot} 
\end{table*}

We then use MvCo-F and MvCo-DoT as baselines to compare with MvCo-CMC and $C^2$M-DoT, respectively, to demonstrate the effectiveness of the cross-modal consistency loss. As shown in Table ~\ref{Table:abalationStudy}, after introducing cross-modal consistency loss, the performance of most natural language metrics of MvCo-CMC and $C^2$M-DoT are further improved, which means that a consistent semantic representation allows the reported text output to be accurate and reliable (closer to the semantics of images).

As shown in Fig.~\ref{fig:multisim}, we visualized the multimodal semantic similarity matrices of frontal image reports and profile image reports of different eras on the IU-X-Ray dataset.The higher the semantic similarity between the report and the image, the lighter the color of the corresponding position in the matrix, otherwise the darker the color. It can be seen that with the increase of training epochs, the graphic-text similarity matrix of the frontal reasoning report and the graphic-text similarity matrix of the lateral reasoning report are more and more similar to the corresponding real report graphic-text similarity matrix. This fully demonstrates the role of our proposed multimodal consistency loss in image-text matching learning. At the same time, the color distinction in the matrix is gradually obvious, indicating that the multimodal consistency loss helps to generate more sample-individualized report results.

\subsection{Additional Results}

\subsubsection{Effects of using multi-view contrastive learning in different positions}
\label{mc}

\begin{figure}[!t]
\begin{minipage}[a]{0.48\linewidth}
  \centering
  \centerline{\includegraphics[width=4.6cm]{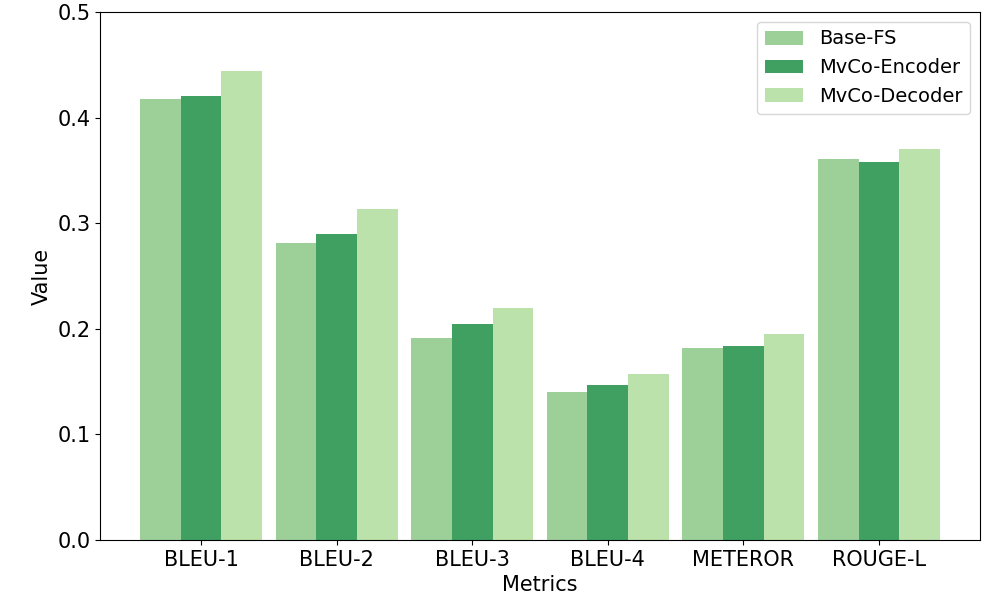}}
  \centerline{(a) IU-X-Ray}\medskip
\end{minipage}
\hfill
\begin{minipage}[a]{0.48\linewidth}
  \centering
  \centerline{\includegraphics[width=4.6cm]{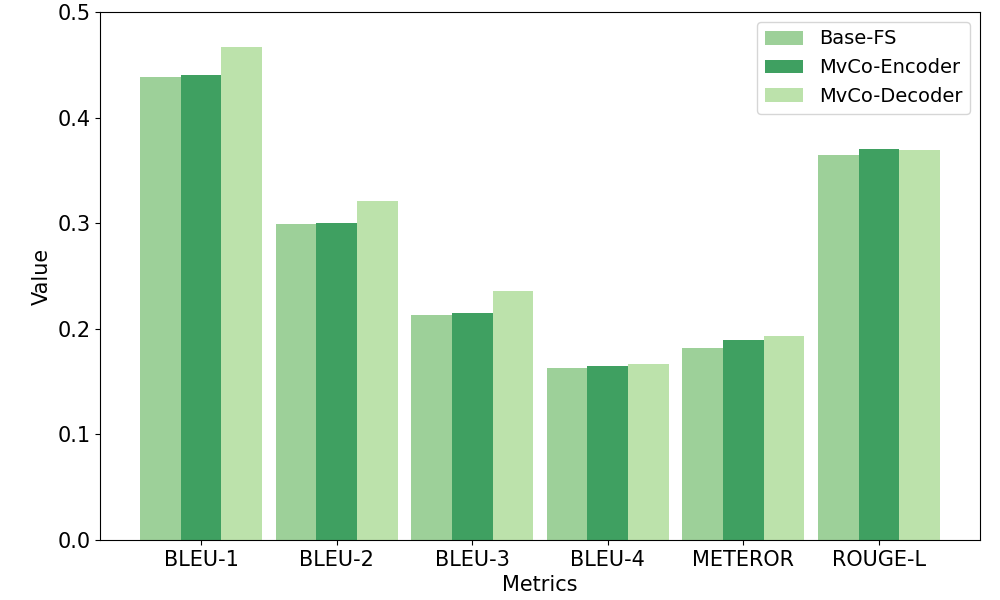}}
  \centerline{(b) MIMIC-CXR}\medskip
\end{minipage}
\vspace{-1em}
\caption{Comparison of the effect of using multi-view contrastive learning in different positions.}
\vspace{-1.5em}
\label{fig.mvco}
\end{figure}
In order to further verify the rationality of multi-view contrastive learning based on semantic features, we compared the impact of using contrastive learning at different locations on the model, as shown in the Fig.~\ref{fig.mvco}. Base-FS represents a fully supervised medical report generation model that does not use multi-view contrastive learning; MvCo-Encoder represents contrastive learning using the feature vectors of the frontal and lateral images output by the encoder; MvCo-Decoder represents using the semantic feature vectors of frontal and lateral reports output by the decoder for comparative learning.

We found that MvCo-Encoder performed better than Base-FS on all metrics, which shows that the mutual information between different views does help focus and understand the lesion characteristics. However, multi-view contrastive learning based on visual features does not significantly improve the performance of report generation. This may be because the spatial features of different views are too different and the visual consistency is limited; in addition, simply updating the encoder to optimize visual feature representation does not help much in end-to-end generation tasks.

When multi-view contrastive learning is used on the decoded semantic vectors, the performance of MvCo-Decoder is further improved. Due to the high degree of agreement between the semantic features decoded from frontal and lateral views of the same patient, more accurate reports were able to be generated. All findings demonstrate that the closer the position using multi-view contrastive learning is to the output, the better the effect. Finally, we use semantic-based multi-view contrastive learning in our research.
\begin{table*}[!t] 
	\caption{
	Results of varying different consistency losses. 
	}
	\centering   
     \resizebox{0.85\textwidth}{!}{
	\begin{tabular}{c|c|cccccc}
		\hline  
		Dataset & Model & BLEU-1 & BLEU-2 & BLEU-3 & BLEU-4  & METEOR & ROUGE-L   \\ \hline 
		\multirow{5}{*}{\scriptsize{\textbf{IU X-Ray}}} 
		& Base-$L_{CE} $  & 0.4175 & 0.2813 & 0.1915 & 0.1400 &0.1820 & 0.3604  \\
		& CMC-$L_{CL} $  & 0.4518 & 0.3091 & 0.2180 & 0.1483  & 0.1872 & 0.3665   \\
		& CMC-$L_{MSE} $  &0.419& 0.2983 & 0.2036 & 0.1361  & 0.1912 & 0.3700  \\
		  & CMC-$L_{JS} $  &0.4452& 0.3131 & 0.2135 & 0.1466  & 0.1944 & 0.3740  \\
            & CMC-$L_{KL} $  & 0.4579 & 0.3214 & 0.2302 & 0.1593 & 0.2037 & 0.3803  \\
		\hline
		\multirow{5}{*}{\scriptsize{\textbf{MIMIC-CXR}}}  
		&   Base-$L_{CE}  $ & 0.4380 & 0.2995 & 0.2132 & 0.1626  & 0.1817 & 0.3647 \\
            &   CMC-$L_{CL}  $   & 0.4633 & 0.3153 & 0.2369 & 0.1901  &0.2008 & 0.3710  \\
		&   CMC-$L_{MSE} $   &0.4672& 0.3205 & 0.2254 &0.1894  & 0.1942 & 0.3640  \\
            &    CMC-$L_{JS} $   &0.4790& 0.3300 & 0.2347 &0.1904 & 0.1922 & 0.3779  \\
            &   CMC-$L_{KL} $   &0.4842 & 0.3450 & 0.2579 &0.1925  & 0.2098 & 0.3850  \\
		
		\hline 
	\end{tabular} }
	
	\vspace*{-1em}
	\label{Table:loss} 
\end{table*}

\subsubsection{Effect of different input sampling methods}

In order to further study the rationality of adaptive input selection methods in domain transfer networks, we compared models using three sampling methods: \emph{random}, \emph{argmax}, and \emph{gumbel}, called DoT-Random, DoT-Argmax, and DoT-Gumbel, respectively. Table~\ref{Table:dot} shows the results of their inference using frontal- or lateral-view, or multi-view inputs.

Specifically, DoT-Argmax achieved the highest scores on BLEU metrics when using multi-view inference, but the performance dropped significantly when using frontal or lateral view inference alone. \emph{Argmax} only samples the input option with the highest probability to participate in training, and the probability value is directly determined by the input information. Since multi-view input tend to contain more feature information (i.e. obtain greater probabilities) than individual frontal- and lateral views, single-view data rarely has a chance to participate in training. The variety of inputs is limited, and thus the improvement on the domain shift problem is limited.

In contrast, the performance of DoT-Random has been further improved when using frontal-view for reasoning. The \emph{random} sampling method ensures the diversity and comprehensiveness of the input, and can effectively alleviate the problem of domain shift. However, due to the inability to select the appropriate input according to the input feature information, the overall reasoning ability obtained is not good.

Finally, we found that DoT-Gumbel achieved the highest scores on almost all metrics when using frontal- and lateral-view reasoning alone, and also achieved the highest METEOR and ROUGE-L results for multi-view reasoning. This is because \emph{gumbel} uses the current input information and adaptively samples based on probability, which can not only select appropriate input features for the model, but also expand the input distribution to a certain extent. Therefore, we adopt \emph{gumbel}-based input sampling in our domain transfer network research.

\subsubsection{Effect of using different consistency losses}

In order to further study the rationality of cross-modal consistency optimization, we compare the effect of using different loss functions. As shown in Table~\ref{Table:loss}, we use the cross-entropy loss-optimized medical report generation method Base-$L_{CE}$ as the baseline, and additionally implement four models of cross-modal loss: (i) CMC-$L_{CL}$: First, the cross-modal semantic similarity matrix is obtained by using the visual features of the image and the semantic features of the prediction report, and then follow the method in CLIP to make it consistent with the diagonal matrix for contrastive learning (CL) (ii) CMC-$L_{MSE}$: Additionally compute a cross-modal semantic similarity matrix between the visual features of an image and the ground-truth reported semantic features, using Mean Squared Error (MSE) for both matrices (iii) CMC-$L_{JS}$: For two cross-modal semantic similarity The matrix uses Jensen-Shannon Divergence (JS divergence) (iv) CMC-$L_{KL}$: Use Kullback-Leibler Divergence (KL divergence) for two cross-modal semantic similarity matrices.

Intuitively, Base-$L_{CE}$ yields the worst results among all losses, which strongly supports our previous theoretical analysis that optimization of unimodality using cross-entropy losses ignores reporting semantics and cannot fully optimized. Therefore, we are motivated to introduce a semantic consistency loss across modalities. It can be seen that CMC-$L_{CL}$ outperforms Base-$L_{CE}$ by a large margin on all six evaluation metrics by using contrastive learning across pairs of images and texts for cross-modal semantics. However, this improvement is limited. Since the images and reports of different patients may have the same semantics, the one-to-one matching mechanism of contrastive learning often lacks the ability to handle one-to-many samples. In contrast, the three models CMC-$L_{MSE}$, CMC-$L_{JS}$ and CMC-$L_{KL}$ that perform consistent calculations on two cross-modal semantic similarity matrices obtain better scores, and since they achieve accurate and flexible cross-modal optimization by realizing the consistent distribution of the semantic similarity matrix of predicted report and vision and the semantic similarity matrix of real report and vision. Specifically, CMC-$L_{KL}$ is more suitable for the medical report generation task and achieves the best performance due to its ability to focus on the fine-grained differences in the distribution rather than the overall similarity of the matrix. Ultimately, we choose to use KL divergence to optimize for cross-modal semantic consistency.



\section{Conclusions}
\label{sec5}
In this paper, to overcome the above problems, we propose a cross-modal consistent multi-view medical report generation with domain transfer network ($C^2$M-DoT). A semantic-based multi-view contrastive learning is proposed to mine the mutual information between different views of medical images to generate more accurate reports; moreover, we also propose to use a domain transfer network based on adaptive input selection to overcome the input distribution gap between the training and inference stages to make the multi-view medical report generation model adaptable to various single-view reasoning. Multimodal optimization based on cross-modal consistency is also used to help text-image matching. We conduct extensive experiments on publicly available datasets IU X-Ray and MIMIC-CXR, demonstrating the superiority and effectiveness of our proposed method.

\section*{Acknowledgments}
This work was supported by the National Natural Science Foundation of China under the grants 62276089, 61906063 and 62102265, by the Natural Science Foundation of Hebei Province, China, under the grant F2021202064, by the ``100 Talents Plan'' of Hebei Province, China, under the grant E2019050017, by the Open Research Fund from Guangdong Laboratory of Artificial Intelligence and Digital Economy (SZ) under the grant GML-KF-22-29, and by the Natural Science Foundation of Guangdong Province of China under the grant 2022A1515011474.

\bibliographystyle{model2-names.bst}\biboptions{authoryear}
\bibliography{refs}

\begin{thebibliography}{46}
\expandafter\ifx\csname natexlab\endcsname\relax\def\natexlab#1{#1}\fi
\providecommand{\url}[1]{\texttt{#1}}
\providecommand{\href}[2]{#2}
\providecommand{\path}[1]{#1}
\providecommand{\DOIprefix}{doi:}
\providecommand{\ArXivprefix}{arXiv:}
\providecommand{\URLprefix}{URL: }
\providecommand{\Pubmedprefix}{pmid:}
\providecommand{\doi}[1]{\href{http://dx.doi.org/#1}{\path{#1}}}
\providecommand{\Pubmed}[1]{\href{pmid:#1}{\path{#1}}}
\providecommand{\bibinfo}[2]{#2}
\ifx\xfnm\relax \def\xfnm[#1]{\unskip,\space#1}\fi
\bibitem[{Amjoud and Amrouch(2021)}]{amjoud2021automatic}
\bibinfo{author}{Amjoud, A.B.}, \bibinfo{author}{Amrouch, M.},
  \bibinfo{year}{2021}.
\newblock \bibinfo{title}{Automatic generation of chest x-ray reports using a
  transformer-based deep learning model}, in: \bibinfo{booktitle}{2021 Fifth
  International Conference on Intelligent Computing in Data Sciences (ICDS)},
  \bibinfo{organization}{IEEE}. pp. \bibinfo{pages}{1--5}.
\bibitem[{Anderson et~al.(2018)Anderson, He, Buehler, Teney, Johnson, Gould and
  Zhang}]{anderson2018bottom}
\bibinfo{author}{Anderson, P.}, \bibinfo{author}{He, X.},
  \bibinfo{author}{Buehler, C.}, \bibinfo{author}{Teney, D.},
  \bibinfo{author}{Johnson, M.}, \bibinfo{author}{Gould, S.},
  \bibinfo{author}{Zhang, L.}, \bibinfo{year}{2018}.
\newblock \bibinfo{title}{Bottom-up and top-down attention for image captioning
  and visual question answering}, in: \bibinfo{booktitle}{Proceedings of CVPR},
  pp. \bibinfo{pages}{6077--6086}.
\bibitem[{Azizi et~al.(2021)Azizi, Mustafa, Ryan, Beaver, Freyberg, Deaton,
  Loh, Karthikesalingam, Kornblith, Chen et~al.}]{azizi2021big}
\bibinfo{author}{Azizi, S.}, \bibinfo{author}{Mustafa, B.},
  \bibinfo{author}{Ryan, F.}, \bibinfo{author}{Beaver, Z.},
  \bibinfo{author}{Freyberg, J.}, \bibinfo{author}{Deaton, J.},
  \bibinfo{author}{Loh, A.}, \bibinfo{author}{Karthikesalingam, A.},
  \bibinfo{author}{Kornblith, S.}, \bibinfo{author}{Chen, T.}, et~al.,
  \bibinfo{year}{2021}.
\newblock \bibinfo{title}{Big self-supervised models advance medical image
  classification}, in: \bibinfo{booktitle}{Proceedings of ICCV}, pp.
  \bibinfo{pages}{3478--3488}.
\bibitem[{Banerjee and Lavie(2005)}]{banerjee2005meteor}
\bibinfo{author}{Banerjee, S.}, \bibinfo{author}{Lavie, A.},
  \bibinfo{year}{2005}.
\newblock \bibinfo{title}{Meteor: An automatic metric for mt evaluation with
  improved correlation with human judgments}, in:
  \bibinfo{booktitle}{Proceedings of ACL}, pp. \bibinfo{pages}{65--72}.
\bibitem[{Chen et~al.(2020a)Chen, Kornblith, Norouzi and
  Hinton}]{chen2020simple}
\bibinfo{author}{Chen, T.}, \bibinfo{author}{Kornblith, S.},
  \bibinfo{author}{Norouzi, M.}, \bibinfo{author}{Hinton, G.},
  \bibinfo{year}{2020}a.
\newblock \bibinfo{title}{A simple framework for contrastive learning of visual
  representations}, in: \bibinfo{booktitle}{Proceedings of ICML}, pp.
  \bibinfo{pages}{1597--1607}.
\bibitem[{Chen et~al.(2020b)Chen, Song, Chang and Wan}]{chen2020generating}
\bibinfo{author}{Chen, Z.}, \bibinfo{author}{Song, Y.}, \bibinfo{author}{Chang,
  T.H.}, \bibinfo{author}{Wan, X.}, \bibinfo{year}{2020}b.
\newblock \bibinfo{title}{Generating radiology reports via memory-driven
  transformer}.
\newblock \bibinfo{journal}{arXiv preprint arXiv:2010.16056} .
\bibitem[{Demner-Fushman et~al.(2016)Demner-Fushman, Kohli, Rosenman, Shooshan,
  Rodriguez, Antani, Thoma and McDonald}]{demner2016preparing}
\bibinfo{author}{Demner-Fushman, D.}, \bibinfo{author}{Kohli, M.D.},
  \bibinfo{author}{Rosenman, M.B.}, \bibinfo{author}{Shooshan, S.E.},
  \bibinfo{author}{Rodriguez, L.}, \bibinfo{author}{Antani, S.},
  \bibinfo{author}{Thoma, G.R.}, \bibinfo{author}{McDonald, C.J.},
  \bibinfo{year}{2016}.
\newblock \bibinfo{title}{Preparing a collection of radiology examinations for
  distribution and retrieval}.
\newblock \bibinfo{journal}{Journal of the American Medical Informatics
  Association} \bibinfo{volume}{23}, \bibinfo{pages}{304--310}.
\bibitem[{Deng et~al.(2009)Deng, Dong, Socher, Li, Li and
  Fei-Fei}]{deng2009imagenet}
\bibinfo{author}{Deng, J.}, \bibinfo{author}{Dong, W.},
  \bibinfo{author}{Socher, R.}, \bibinfo{author}{Li, L.J.},
  \bibinfo{author}{Li, K.}, \bibinfo{author}{Fei-Fei, L.},
  \bibinfo{year}{2009}.
\newblock \bibinfo{title}{Imagenet: A large-scale hierarchical image database},
  in: \bibinfo{booktitle}{Proceedings of CVPR}, pp. \bibinfo{pages}{248--255}.
\bibitem[{Dosovitskiy et~al.(2020)Dosovitskiy, Beyer, Kolesnikov, Weissenborn,
  Zhai, Unterthiner, Dehghani, Minderer, Heigold, Gelly
  et~al.}]{dosovitskiy2020image}
\bibinfo{author}{Dosovitskiy, A.}, \bibinfo{author}{Beyer, L.},
  \bibinfo{author}{Kolesnikov, A.}, \bibinfo{author}{Weissenborn, D.},
  \bibinfo{author}{Zhai, X.}, \bibinfo{author}{Unterthiner, T.},
  \bibinfo{author}{Dehghani, M.}, \bibinfo{author}{Minderer, M.},
  \bibinfo{author}{Heigold, G.}, \bibinfo{author}{Gelly, S.}, et~al.,
  \bibinfo{year}{2020}.
\newblock \bibinfo{title}{An image is worth 16x16 words: Transformers for image
  recognition at scale}.
\newblock \bibinfo{journal}{arXiv preprint arXiv:2010.11929} .
\bibitem[{Eslami et~al.(2023)Eslami, Meinel and
  de~Melo}]{eslami-etal-2023-pubmedclip}
\bibinfo{author}{Eslami, S.}, \bibinfo{author}{Meinel, C.},
  \bibinfo{author}{de~Melo, G.}, \bibinfo{year}{2023}.
\newblock \bibinfo{title}{{P}ub{M}ed{CLIP}: How much does {CLIP} benefit visual
  question answering in the medical domain?}, in:
  \bibinfo{booktitle}{Proceedings of CVPR}, pp. \bibinfo{pages}{1181--1193}.
\bibitem[{Gao et~al.(2021)Gao, Yao and Chen}]{gao2021simcse}
\bibinfo{author}{Gao, T.}, \bibinfo{author}{Yao, X.}, \bibinfo{author}{Chen,
  D.}, \bibinfo{year}{2021}.
\newblock \bibinfo{title}{Simcse: Simple contrastive learning of sentence
  embeddings}.
\newblock \bibinfo{journal}{arXiv preprint arXiv:2104.08821} .
\bibitem[{He et~al.(2020)He, Fan, Wu, Xie and Girshick}]{he2020momentum}
\bibinfo{author}{He, K.}, \bibinfo{author}{Fan, H.}, \bibinfo{author}{Wu, Y.},
  \bibinfo{author}{Xie, S.}, \bibinfo{author}{Girshick, R.},
  \bibinfo{year}{2020}.
\newblock \bibinfo{title}{Momentum contrast for unsupervised visual
  representation learning}, in: \bibinfo{booktitle}{Proceedings of CVPR}, pp.
  \bibinfo{pages}{9729--9738}.
\bibitem[{He et~al.(2016)He, Zhang, Ren and Sun}]{he2016deep}
\bibinfo{author}{He, K.}, \bibinfo{author}{Zhang, X.}, \bibinfo{author}{Ren,
  S.}, \bibinfo{author}{Sun, J.}, \bibinfo{year}{2016}.
\newblock \bibinfo{title}{Deep residual learning for image recognition}, in:
  \bibinfo{booktitle}{Proceedings of CVPR}, pp. \bibinfo{pages}{770--778}.
\bibitem[{Hou et~al.(2021)Hou, Kaissis, Summers and Kainz}]{hou2021ratchet}
\bibinfo{author}{Hou, B.}, \bibinfo{author}{Kaissis, G.},
  \bibinfo{author}{Summers, R.M.}, \bibinfo{author}{Kainz, B.},
  \bibinfo{year}{2021}.
\newblock \bibinfo{title}{Ratchet: Medical transformer for chest x-ray
  diagnosis and reporting}, in: \bibinfo{booktitle}{Proceedings of MICCAI}, pp.
  \bibinfo{pages}{293--303}.
\bibitem[{Huang et~al.(2021)Huang, Shen, Lungren and Yeung}]{huang2021gloria}
\bibinfo{author}{Huang, S.C.}, \bibinfo{author}{Shen, L.},
  \bibinfo{author}{Lungren, M.P.}, \bibinfo{author}{Yeung, S.},
  \bibinfo{year}{2021}.
\newblock \bibinfo{title}{Gloria: A multimodal global-local representation
  learning framework for label-efficient medical image recognition}, in:
  \bibinfo{booktitle}{Proceedings of the IEEE/CVF International Conference on
  Computer Vision（ICCV）}, pp. \bibinfo{pages}{3942--3951}.
\bibitem[{Jang et~al.(2016)Jang, Gu and Poole}]{jang2016categorical}
\bibinfo{author}{Jang, E.}, \bibinfo{author}{Gu, S.}, \bibinfo{author}{Poole,
  B.}, \bibinfo{year}{2016}.
\newblock \bibinfo{title}{Categorical reparameterization with gumbel-softmax}.
\newblock \bibinfo{journal}{arXiv preprint arXiv:1611.01144} .
\bibitem[{Jing et~al.(2019)Jing, Wang and Xing}]{jing-etal-2019-show}
\bibinfo{author}{Jing, B.}, \bibinfo{author}{Wang, Z.}, \bibinfo{author}{Xing,
  E.}, \bibinfo{year}{2019}.
\newblock \bibinfo{title}{Show, describe and conclude: On exploiting the
  structure information of chest {X}-ray reports}, in:
  \bibinfo{booktitle}{Proceedings of the Association for Computational
  Linguistics}, pp. \bibinfo{pages}{6570--6580}.
\newblock \DOIprefix\doi{10.18653/v1/P19-1657}.
\bibitem[{Jing et~al.(2018)Jing, Xie and Xing}]{jing-etal-2018-automatic}
\bibinfo{author}{Jing, B.}, \bibinfo{author}{Xie, P.}, \bibinfo{author}{Xing,
  E.}, \bibinfo{year}{2018}.
\newblock \bibinfo{title}{On the automatic generation of medical imaging
  reports}, in: \bibinfo{booktitle}{Proceedings of the Association for
  Computational Linguistics (Volume 1: Long Papers)}, pp.
  \bibinfo{pages}{2577--2586}.
\bibitem[{Johnson et~al.(2019)Johnson, Pollard, Berkowitz, Greenbaum, Lungren,
  Deng, Mark and Horng}]{johnson2019mimic}
\bibinfo{author}{Johnson, A.E.}, \bibinfo{author}{Pollard, T.J.},
  \bibinfo{author}{Berkowitz, S.J.}, \bibinfo{author}{Greenbaum, N.R.},
  \bibinfo{author}{Lungren, M.P.}, \bibinfo{author}{Deng, C.y.},
  \bibinfo{author}{Mark, R.G.}, \bibinfo{author}{Horng, S.},
  \bibinfo{year}{2019}.
\newblock \bibinfo{title}{{MIMIC-CXR}, a de-identified publicly available
  database of chest radiographs with free-text reports}.
\newblock \bibinfo{journal}{Sci. Data} \bibinfo{volume}{6},
  \bibinfo{pages}{317}.
\newblock
  \bibinfo{note}{Doi:{\href{https://doi.org/10.1038/s41597-019-0322-0}{10.1038/s41597-019-0322-0}}}.
\bibitem[{Kingma and Ba(2014)}]{kingma2014adam}
\bibinfo{author}{Kingma, D.P.}, \bibinfo{author}{Ba, J.}, \bibinfo{year}{2014}.
\newblock \bibinfo{title}{Adam: A method for stochastic optimization}.
\newblock \bibinfo{journal}{arXiv preprint arXiv:1412.6980} .
\bibitem[{Li et~al.(2018)Li, Liang, Hu and Xing}]{li2018hybrid}
\bibinfo{author}{Li, Y.}, \bibinfo{author}{Liang, X.}, \bibinfo{author}{Hu,
  Z.}, \bibinfo{author}{Xing, E.P.}, \bibinfo{year}{2018}.
\newblock \bibinfo{title}{Hybrid retrieval-generation reinforced agent for
  medical image report generation}, in: \bibinfo{booktitle}{Advances in Neural
  Information Processing Systems}, pp. \bibinfo{pages}{1530--1540}.
\bibitem[{Lin(2004)}]{lin2004rouge}
\bibinfo{author}{Lin, C.Y.}, \bibinfo{year}{2004}.
\newblock \bibinfo{title}{Rouge: A package for automatic evaluation of
  summaries}, in: \bibinfo{booktitle}{Proceedings of ACL}, pp.
  \bibinfo{pages}{74--81}.
\bibitem[{Liu et~al.(2021)Liu, Wu, Ge, Fan and Zou}]{liu2021exploring}
\bibinfo{author}{Liu, F.}, \bibinfo{author}{Wu, X.}, \bibinfo{author}{Ge, S.},
  \bibinfo{author}{Fan, W.}, \bibinfo{author}{Zou, Y.}, \bibinfo{year}{2021}.
\newblock \bibinfo{title}{Exploring and distilling posterior and prior
  knowledge for radiology report generation}, in:
  \bibinfo{booktitle}{Proceedings of CVPR}, pp. \bibinfo{pages}{13753--13762}.
\bibitem[{Liu et~al.(2019)Liu, Hsu, McDermott, Boag, Weng, Szolovits and
  Ghassemi}]{liu2019clinically}
\bibinfo{author}{Liu, G.}, \bibinfo{author}{Hsu, T.M.H.},
  \bibinfo{author}{McDermott, M.}, \bibinfo{author}{Boag, W.},
  \bibinfo{author}{Weng, W.H.}, \bibinfo{author}{Szolovits, P.},
  \bibinfo{author}{Ghassemi, M.}, \bibinfo{year}{2019}.
\newblock \bibinfo{title}{Clinically accurate chest {X}-ray report generation}.
\newblock \bibinfo{journal}{arXiv preprint arXiv:1904.02633} .
\bibitem[{Nair and Hinton(2010)}]{nair2010rectified}
\bibinfo{author}{Nair, V.}, \bibinfo{author}{Hinton, G.E.},
  \bibinfo{year}{2010}.
\newblock \bibinfo{title}{Rectified linear units improve restricted boltzmann
  machines}, in: \bibinfo{booktitle}{Proceedings of ICML}, pp.
  \bibinfo{pages}{807--814}.
\bibitem[{Pan et~al.(2020)Pan, Yao, Li and Mei}]{pan2020x}
\bibinfo{author}{Pan, Y.}, \bibinfo{author}{Yao, T.}, \bibinfo{author}{Li, Y.},
  \bibinfo{author}{Mei, T.}, \bibinfo{year}{2020}.
\newblock \bibinfo{title}{X-linear attention networks for image captioning},
  in: \bibinfo{booktitle}{Proceedings of CVPR}, pp.
  \bibinfo{pages}{10971--10980}.
\bibitem[{Papineni et~al.(2002)Papineni, Roukos, Ward and
  Zhu}]{papineni2002bleu}
\bibinfo{author}{Papineni, K.}, \bibinfo{author}{Roukos, S.},
  \bibinfo{author}{Ward, T.}, \bibinfo{author}{Zhu, W.J.},
  \bibinfo{year}{2002}.
\newblock \bibinfo{title}{Bleu: a method for automatic evaluation of machine
  translation}, in: \bibinfo{booktitle}{Proceedings of ACL}, pp.
  \bibinfo{pages}{311--318}.
\bibitem[{Pelka et~al.(2018)Pelka, Koitka, R{\"u}ckert, Nensa and
  Friedrich}]{pelka2018radiology}
\bibinfo{author}{Pelka, O.}, \bibinfo{author}{Koitka, S.},
  \bibinfo{author}{R{\"u}ckert, J.}, \bibinfo{author}{Nensa, F.},
  \bibinfo{author}{Friedrich, C.M.}, \bibinfo{year}{2018}.
\newblock \bibinfo{title}{Radiology objects in context (roco): a multimodal
  image dataset}, in: \bibinfo{booktitle}{Intravascular Imaging and Computer
  Assisted Stenting and Large-Scale Annotation of Biomedical Data and Expert
  Label Synthesis: 7th Joint International Workshop, CVII-STENT 2018 and Third
  International Workshop, LABELS 2018, Held in Conjunction with MICCAI 2018,
  Granada, Spain, September 16, 2018, Proceedings 3},
  \bibinfo{organization}{Springer}. pp. \bibinfo{pages}{180--189}.
\bibitem[{Radford et~al.(2021)Radford, Kim, Hallacy, Ramesh, Goh, Agarwal,
  Sastry, Askell, Mishkin, Clark et~al.}]{radford2021learning}
\bibinfo{author}{Radford, A.}, \bibinfo{author}{Kim, J.W.},
  \bibinfo{author}{Hallacy, C.}, \bibinfo{author}{Ramesh, A.},
  \bibinfo{author}{Goh, G.}, \bibinfo{author}{Agarwal, S.},
  \bibinfo{author}{Sastry, G.}, \bibinfo{author}{Askell, A.},
  \bibinfo{author}{Mishkin, P.}, \bibinfo{author}{Clark, J.}, et~al.,
  \bibinfo{year}{2021}.
\newblock \bibinfo{title}{Learning transferable visual models from natural
  language supervision}, in: \bibinfo{booktitle}{Proceedings of International
  conference on machine learning（ICML）}, pp. \bibinfo{pages}{8748--8763}.
\bibitem[{Rennie et~al.(2017)Rennie, Marcheret, Mroueh, Ross and
  Goel}]{rennie2017self}
\bibinfo{author}{Rennie, S.J.}, \bibinfo{author}{Marcheret, E.},
  \bibinfo{author}{Mroueh, Y.}, \bibinfo{author}{Ross, J.},
  \bibinfo{author}{Goel, V.}, \bibinfo{year}{2017}.
\newblock \bibinfo{title}{Self-critical sequence training for image
  captioning}, in: \bibinfo{booktitle}{Proceedings of CVPR}, pp.
  \bibinfo{pages}{7008--7024}.
\bibitem[{Seibold et~al.(2022)Seibold, Rei{\ss}, Sarfraz, Stiefelhagen and
  Kleesiek}]{seibold2022breaking}
\bibinfo{author}{Seibold, C.}, \bibinfo{author}{Rei{\ss}, S.},
  \bibinfo{author}{Sarfraz, M.S.}, \bibinfo{author}{Stiefelhagen, R.},
  \bibinfo{author}{Kleesiek, J.}, \bibinfo{year}{2022}.
\newblock \bibinfo{title}{Breaking with fixed set pathology recognition through
  report-guided contrastive training}, in: \bibinfo{booktitle}{Proceedings of
  International Conference on Medical Image Computing and Computer-Assisted
  Intervention（MICCAI）}, pp. \bibinfo{pages}{690--700}.
\bibitem[{Tian et~al.(2020)Tian, Krishnan and Isola}]{tian2020contrastive}
\bibinfo{author}{Tian, Y.}, \bibinfo{author}{Krishnan, D.},
  \bibinfo{author}{Isola, P.}, \bibinfo{year}{2020}.
\newblock \bibinfo{title}{Contrastive multiview coding}, in:
  \bibinfo{booktitle}{Proceedings of ECCV}, pp. \bibinfo{pages}{776--794}.
\bibitem[{Vijayakumar et~al.(2016)Vijayakumar, Cogswell, Selvaraju, Sun, Lee,
  Crandall and Batra}]{vijayakumar2016diverse}
\bibinfo{author}{Vijayakumar, A.K.}, \bibinfo{author}{Cogswell, M.},
  \bibinfo{author}{Selvaraju, R.R.}, \bibinfo{author}{Sun, Q.},
  \bibinfo{author}{Lee, S.}, \bibinfo{author}{Crandall, D.},
  \bibinfo{author}{Batra, D.}, \bibinfo{year}{2016}.
\newblock \bibinfo{title}{Diverse beam search: Decoding diverse solutions from
  neural sequence models}.
\newblock \bibinfo{journal}{arXiv preprint arXiv:1610.02424} .
\bibitem[{Vu et~al.(2021)Vu, Wang, Balachandar, Liu, Ng and
  Rajpurkar}]{vu2021medaug}
\bibinfo{author}{Vu, Y.N.T.}, \bibinfo{author}{Wang, R.},
  \bibinfo{author}{Balachandar, N.}, \bibinfo{author}{Liu, C.},
  \bibinfo{author}{Ng, A.Y.}, \bibinfo{author}{Rajpurkar, P.},
  \bibinfo{year}{2021}.
\newblock \bibinfo{title}{Medaug: Contrastive learning leveraging patient
  metadata improves representations for chest x-ray interpretation}, in:
  \bibinfo{booktitle}{Proceedings of MLHC}, pp. \bibinfo{pages}{755--769}.
\bibitem[{Wang et~al.(2022)Wang, Wu, Agarwal and Sun}]{wang2022medclip}
\bibinfo{author}{Wang, Z.}, \bibinfo{author}{Wu, Z.}, \bibinfo{author}{Agarwal,
  D.}, \bibinfo{author}{Sun, J.}, \bibinfo{year}{2022}.
\newblock \bibinfo{title}{Medclip: Contrastive learning from unpaired medical
  images and text}.
\newblock \bibinfo{journal}{arXiv preprint arXiv:2210.10163} .
\bibitem[{Wu et~al.(2023)Wu, Zhang, Zhang, Wang and Xie}]{wu2023medklip}
\bibinfo{author}{Wu, C.}, \bibinfo{author}{Zhang, X.}, \bibinfo{author}{Zhang,
  Y.}, \bibinfo{author}{Wang, Y.}, \bibinfo{author}{Xie, W.},
  \bibinfo{year}{2023}.
\newblock \bibinfo{title}{Medklip: Medical knowledge enhanced language-image
  pre-training}.
\newblock \bibinfo{journal}{medRxiv} , \bibinfo{pages}{2023--01}.
\bibitem[{Xiong et~al.(2019)Xiong, Du and Yan}]{10.1007/978-3-030-32692-0_77}
\bibinfo{author}{Xiong, Y.}, \bibinfo{author}{Du, B.}, \bibinfo{author}{Yan,
  P.}, \bibinfo{year}{2019}.
\newblock \bibinfo{title}{Reinforced transformer for medical image captioning},
  in: \bibinfo{booktitle}{Proceedings of MICCAI}, pp.
  \bibinfo{pages}{673--680}.
\bibitem[{Xu et~al.(2022)Xu, Xu, Chen, Qi and Lukasiewicz}]{xu2020reinforced}
\bibinfo{author}{Xu, W.}, \bibinfo{author}{Xu, Z.}, \bibinfo{author}{Chen, J.},
  \bibinfo{author}{Qi, C.}, \bibinfo{author}{Lukasiewicz, T.},
  \bibinfo{year}{2022}.
\newblock \bibinfo{title}{Hybrid reinforced medical report generation with
  m-linear attention and repetition penalty}.
\newblock \bibinfo{journal}{arXiv preprint arXiv:2210.13729} .
\bibitem[{Xue et~al.(2018)Xue, Xu, Rodney~Long, Xue, Antani, Thoma and
  Huang}]{xue2018multimodal}
\bibinfo{author}{Xue, Y.}, \bibinfo{author}{Xu, T.},
  \bibinfo{author}{Rodney~Long, L.}, \bibinfo{author}{Xue, Z.},
  \bibinfo{author}{Antani, S.}, \bibinfo{author}{Thoma, G.R.},
  \bibinfo{author}{Huang, X.}, \bibinfo{year}{2018}.
\newblock \bibinfo{title}{Multimodal recurrent model with attention for
  automated radiology report generation}, in: \bibinfo{booktitle}{Proceedings
  of MICCAI}, pp. \bibinfo{pages}{457--466}.
\bibitem[{Yan et~al.(2021)Yan, He, Lu, Du, Chang, Gentili, McAuley and
  Hsu}]{yan2021weakly}
\bibinfo{author}{Yan, A.}, \bibinfo{author}{He, Z.}, \bibinfo{author}{Lu, X.},
  \bibinfo{author}{Du, J.}, \bibinfo{author}{Chang, E.},
  \bibinfo{author}{Gentili, A.}, \bibinfo{author}{McAuley, J.},
  \bibinfo{author}{Hsu, C.N.}, \bibinfo{year}{2021}.
\newblock \bibinfo{title}{Weakly supervised contrastive learning for chest
  x-ray report generation}.
\newblock \bibinfo{journal}{arXiv preprint arXiv:2109.12242} .
\bibitem[{You et~al.(2021)You, Liu, Ge, Xie, Zhang and
  Wu}]{you2021aligntransformer}
\bibinfo{author}{You, D.}, \bibinfo{author}{Liu, F.}, \bibinfo{author}{Ge, S.},
  \bibinfo{author}{Xie, X.}, \bibinfo{author}{Zhang, J.}, \bibinfo{author}{Wu,
  X.}, \bibinfo{year}{2021}.
\newblock \bibinfo{title}{Aligntransformer: Hierarchical alignment of visual
  regions and disease tags for medical report generation}, in:
  \bibinfo{booktitle}{Proceedings of Medical Image Computing and Computer
  Assisted Intervention（MICCAI）}, pp. \bibinfo{pages}{72--82}.
\bibitem[{Yuan et~al.(2019)Yuan, Liao, Luo and Luo}]{yuan2019automatic}
\bibinfo{author}{Yuan, J.}, \bibinfo{author}{Liao, H.}, \bibinfo{author}{Luo,
  R.}, \bibinfo{author}{Luo, J.}, \bibinfo{year}{2019}.
\newblock \bibinfo{title}{Automatic radiology report generation based on
  multi-view image fusion and medical concept enrichment}, in:
  \bibinfo{booktitle}{Medical Image Computing and Computer Assisted
  Intervention--MICCAI 2019: 22nd International Conference, Shenzhen, China,
  October 13--17, 2019, Proceedings, Part VI 22}, pp.
  \bibinfo{pages}{721--729}.
\bibitem[{Zhang et~al.(2023a)Zhang, Zhang, Shen, Lukasiewicz and
  Xu}]{zhang2023multi}
\bibinfo{author}{Zhang, J.}, \bibinfo{author}{Zhang, S.},
  \bibinfo{author}{Shen, X.}, \bibinfo{author}{Lukasiewicz, T.},
  \bibinfo{author}{Xu, Z.}, \bibinfo{year}{2023}a.
\newblock \bibinfo{title}{Multi-condos: Multimodal contrastive domain sharing
  generative adversarial networks for self-supervised medical image
  segmentation}.
\newblock \bibinfo{journal}{IEEE Transactions on Medical Imaging} .
\bibitem[{Zhang et~al.(2023b)Zhang, Zhang, Tian, Lukasiewicz and
  Xu}]{ZHANG-MIA2022}
\bibinfo{author}{Zhang, S.}, \bibinfo{author}{Zhang, J.},
  \bibinfo{author}{Tian, B.}, \bibinfo{author}{Lukasiewicz, T.},
  \bibinfo{author}{Xu, Z.}, \bibinfo{year}{2023}b.
\newblock \bibinfo{title}{Multi-modal contrastive mutual learning and
  pseudo-label re-learning for semi-supervised medical image segmentation}.
\newblock \bibinfo{journal}{Medical Image Analysis} \bibinfo{volume}{83},
  \bibinfo{pages}{102656}.
\bibitem[{Zhang et~al.(2020)Zhang, Jiang, Miura, Manning and
  Langlotz}]{zhang2020contrastive}
\bibinfo{author}{Zhang, Y.}, \bibinfo{author}{Jiang, H.},
  \bibinfo{author}{Miura, Y.}, \bibinfo{author}{Manning, C.D.},
  \bibinfo{author}{Langlotz, C.P.}, \bibinfo{year}{2020}.
\newblock \bibinfo{title}{Contrastive learning of medical visual
  representations from paired images and text}.
\newblock \bibinfo{journal}{arXiv preprint arXiv:2010.00747} .
\bibitem[{Zhang et~al.(2022)Zhang, Jiang, Miura, Manning and
  Langlotz}]{zhang2022contrastive}
\bibinfo{author}{Zhang, Y.}, \bibinfo{author}{Jiang, H.},
  \bibinfo{author}{Miura, Y.}, \bibinfo{author}{Manning, C.D.},
  \bibinfo{author}{Langlotz, C.P.}, \bibinfo{year}{2022}.
\newblock \bibinfo{title}{Contrastive learning of medical visual
  representations from paired images and text}, in:
  \bibinfo{booktitle}{Proceedings of MLHC}, pp. \bibinfo{pages}{2--25}.

\end{thebibliography}

\end{document}